\documentclass[3p]{elsarticle}
\usepackage{lineno,hyperref}
\modulolinenumbers[5]

\journal{Latex}

\usepackage{amsmath,amsthm}

\usepackage{graphicx}
\usepackage{algpseudocode}
\usepackage{algorithm}
\usepackage{algpascal}
\usepackage{bm}
\usepackage{hyperref}
\usepackage{amssymb}
\usepackage{xcolor}

\DeclareRobustCommand{\uvec}[1]{{%
		\ifcsname uvec#1\endcsname
		\csname uvec#1\endcsname
		\else
		\bm{\mathbf{#1}}%
		\fi
}}

\theoremstyle{plain}
\newtheorem{thm}{Theorem}[]

\newtheorem{prop}{Proposition}

\theoremstyle{definition}

\theoremstyle{remark}
\newtheorem{rem}{Remark}[]

 \DeclareMathOperator{\sign}{sgn}

\begin{document}

\begin{frontmatter}

\title{Motion Planning for a Spin-Rolling Sphere on a Plane}


\author[mymainaddress]{Seyed Amir Tafrishi\corref{mycorrespondingauthor}}
\cortext[mycorrespondingauthor]{Corresponding author}
\ead{s.a.tafrishi@srd.mech.tohoku.ac.jp}

\author[Svinin]{Mikhail Svinin}

\author[Yamamo]{Motoji Yamamoto}

\author[mymainaddress]{Yasuhisa Hirata}

\address[mymainaddress]{Department of Robotics, Tohoku University, Sendai, Japan}
\address[Svinin]{Department of Information Science and Engineering, Ritsumeikan University, Kyoto, Japan}
\address[Yamamo]{Department of Mechanical Engineering, Kyushu University, Kyushu, Japan}

\begin{abstract}
The paper deals with motion planning for a spin-rolling sphere when the sphere follows a straight path on a plane. Since the motion of the sphere is constrained by the straight line, the control of the sphere's spin motion is essential to converge to a desired configuration of the sphere. In this paper, we show a new geometric-based planning approach that is based on a full-state description of this nonlinear system. First, the problem statement of the motion planning is posed. Next, we develop a geometric controller implemented as a virtual surface by using the Darboux frame kinematics. This virtual surface generates arc-length-based inputs for controlling the trajectories of the sphere. Then, an iterative algorithm is designed to tune these inputs for the desired configurations. The feasibility of the proposed approach is verified by simulations.
\end{abstract}


\end{frontmatter}


\section{Introduction}

\begin{figure}[t!]
	\centering
	\includegraphics[width=3.9 in]{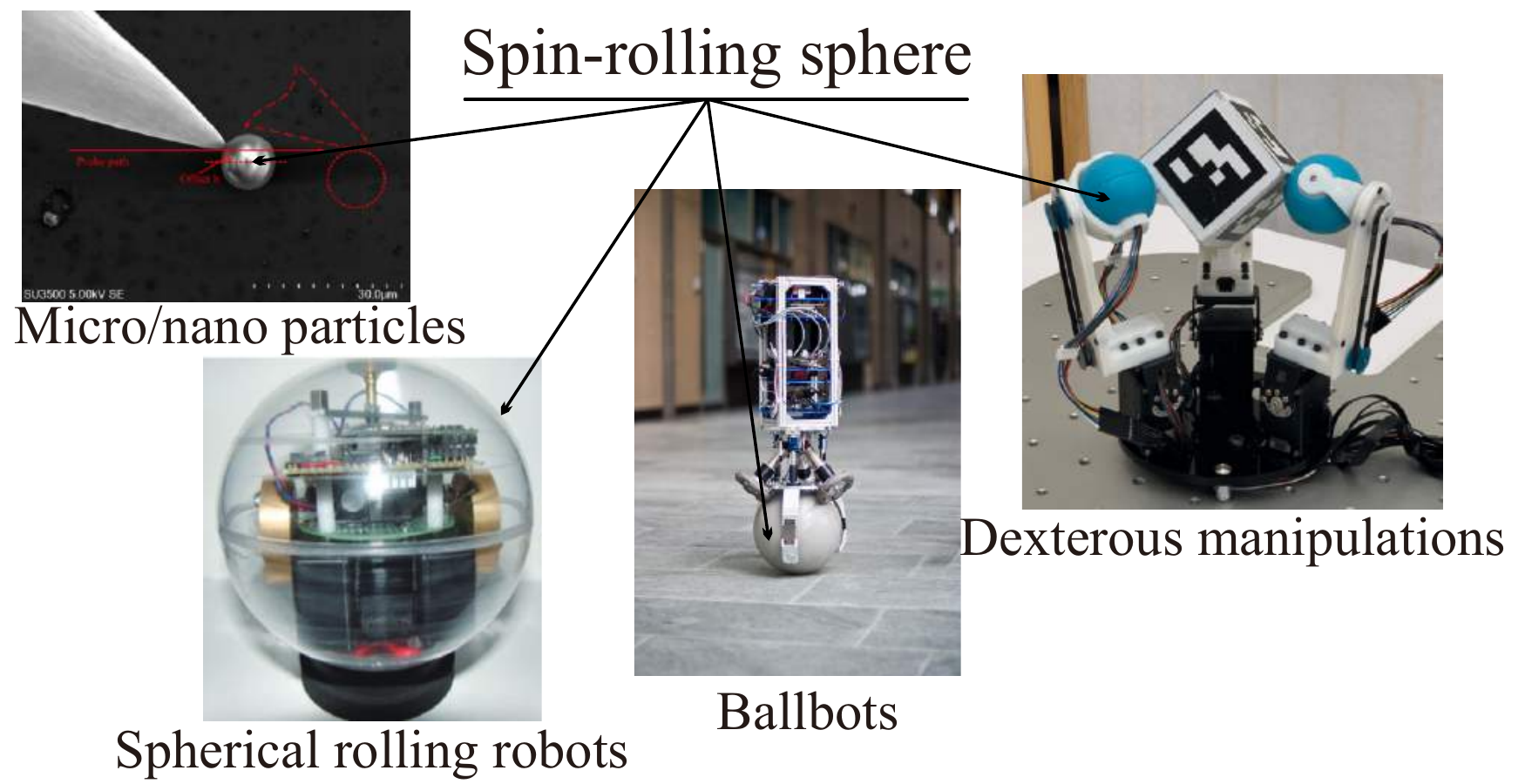}		
	\caption{Realization of the spin-rolling motion of the sphere, e.g., the micro/nano particle manipulation  \cite{sumer2008rolling}, the spherical rolling robots \cite{ishikawa2011volvot}, ballbots \cite{spinjohnson2018fuzzy} and the Dexterous manipulation \cite{SpinrollMechIROS2020}. }\label{Fig:Applicationssphere}
\end{figure}

Manipulation of a spin-rolling sphere to a desired configuration is an important research problem that can find many applications.
Different mechanisms can realize a spin-rolling motion of the sphere on a plane (see Fig.~\ref{Fig:Applicationssphere}). In the field of dexterous manipulation, rolling fingertips with convex surfaces can be utilized to grasp and manipulate different objects \cite{kiss2002motion,cui_sun_dai_2017,SpinrollMechIROS2020}. For example, Yuan et al. developed rolling spherical fingertips for grasping mechanisms \cite{SpinrollMechIROS2020}. There are propulsion systems that can realize spin-rolling motions with multiple-degrees-of-freedom (DoFs) actuators. For instance, spherical mobile robots with mass-imbalance propulsion \cite{RollRollerRobotThesis2014,Tafrishi2019}, with cart-based actuators inside their spherical shell \cite{ishikawa2011volvot,karavaev2020spherical} and internal motor drivers \cite{borisov2012control,borisov2013control,svinin2013dynamic} can have a similar feature with simple increase in number of actuators. Additionally, the ballbot robots with drivers on top of the rolling ball can be included in the kinematic model of spin-rolling motion \cite{fankhauser2010modeling,spinjohnson2018fuzzy}. Furthermore, nano/micro spherical particle manipulation on plane surfaces is finding important applications \cite{sumer2008rolling,diller2013micro,fernandez2017three}. Thus, the problem of path planning for a spin-rolling sphere is applicable to different mechanisms.

The analysis of the spin-rolling motion can be done by using the conventional ball-plate system kinematics \cite{jurdjevic1993geometry,marigo2000rolling} but we cannot spin the ball directly by the rotation of the upper sandwiched plane \cite{Tafrishi2021DarKiF}.
Different kinematic parametrization of rolling contact have been considered in \cite{Tafrishi2021DarKiF,sankar1996velocity,Montana1988,CuiDarboux2010,bizyaev2019different,woodruff2019second}. To conduct motion planning, Kiss et al. \cite{kiss2002motion} used three independent planes to manipulate the sphere at the kinematic level and controlled the relative angles of the sphere without considering desired plane configuration. Date et al. used the advantage of spinning indirectly to control the ball-plate system \cite{date2004simultaneous}. Their motion planning algorithm was based on shifting the coordinate of the actuating plane in iterations with respect to different reference frames. This change can be looked like a rotational virtual center. However, the time scaling with the included coordinate transformation of the kinematic model can bring the system to an uncontrollable state \cite{Oriolo2005Feedback}.

In the literature on the pure rolling, different feed-forward \cite{jurdjevic1993geometry,das2004exponential,alouges2010motion,svinin2013dynamic} and feedback control \cite{date2004simultaneous,Oriolo2005Feedback,mukherjee2002feedback,woodruff2020motion} methods were developed. Since the change of the spin angle affects the orientation of the sphere, the conventional planning approaches, such as the ones based on the geometric phase shifting \cite{planningli1990,mukherjee2002motion,svinin2008motion,kilin2015spherical,bai2018dynamic}, cannot be adjusted naturally for the spin-rolling motion planning. Arthurs et al. \cite{arthurs1986hammersley} and Jurdjevic \cite{jurdjevic1993geometry} proposed planning approaches based on the optimal control theory. This research direction was developed later on by Sachkov \cite{sachkov2010,mashtakov2011}.
As an alternative to feedforward planning, to stabilize the non-differentially flat ball-plate system, an iterative steering was formulated by Oriolo and Vendittelli\cite{Oriolo2005Feedback}. However, the solutions had serious fluctuations as trajectories converge to the desired states. Thus, their created trajectories were not easy to be realized by dynamical systems \cite{morinaga2014motion} and there were singularities in different regions of spherical manifold due to the locality of the solutions.
Beschastnyi \cite{beschatnyi2014optimal} studied the optimality problem of the spin-rolling sphere. Extremal trajectories were parameterized, and their cut times were estimated for optimality. It was proposed that Maxwell time can be determined while the sphere follows the straight path for the optimal solution.
However, this research did not cover the control problem for arbitrary desired states.


In this paper, we approach the motion planning problem with two main motivations: considering a spin-rolling sphere on the plane and developing optimal smooth trajectories for the arbitrary desired states. Our approach is based on parameterizing the spin-rolling sphere motion by using the Darboux-frame-based kinematics of the ball-plate system originally proposed by Cui and Dai \cite{CuiDarboux2010,cui_sun_dai_2017,cui2020sliding}.
In our previous work\cite{Tafrishi2021DarKiF}, we transformed the underactuated ball-plate system to a fully-actuated one with the arc-length-based control inputs. In this paper, after explaining the motion planning problem, we introduce a virtual surface to manipulate the control inputs of the Darboux-frame-based kinematic model to a desired configuration. To the best of our knowledge, this is the first proposed geometric arc-length-based control strategy for the motion planning of a spin-rolling sphere. This control strategy separates the time scale from the kinematic equations, which allows constructing motion with different convergence rates in a given time. Finally, we propose a tuning algorithm to iteratively solve the differential kinematic model until achieving a successful final full configuration. It is important to note that the designed Darboux-frame parametrization increases the number of planning parameters that simplify the developed algorithm.

This paper is organized as follows. In Section 2, the motion planning problem is stated and a distance constraint due to the no-sliding constraint is explained. Next, Section 3 includes the design of the geometric controller for a Darboux frame kinematics. In Section 4, the iterative algorithm with included tuning variables is described. Simulations of this new approach are demonstrated and discussed in Section 5. Finally, we conclude our findings in Section 6.

\section{Problem Statement}
\label{Sec:Planning}
It is assumed that the sphere that rolls and spins along a straight path on the plane, and the
the path on the sphere, connecting the initial and final configurations, must be established.
Because of the no-sliding constraint the length of the curve on the sphere $\uvec{L}_o$ is equal to that on the the plane $\uvec{L}_s$, the curve may not be long enough to reach certain configurations on the sphere. This limitation is mainly due to our considered approach with a restricted path on the plane. To deal with this issue, we find a distance constraint that determines the minimum length of the curve for reaching the sphere to its desired states. 
Note that this limitation only appears when the length between the initial and final states of the plane is less than the circumference of the sphere. We calculate this distance constraint for the plane by applying the Gauss-Bonnet theorem \cite{DIfgeometry1976} to the created circular path on the desired angular states of the sphere.

\begin{figure}[t!]
	\centering
	\includegraphics[width=3.9 in]{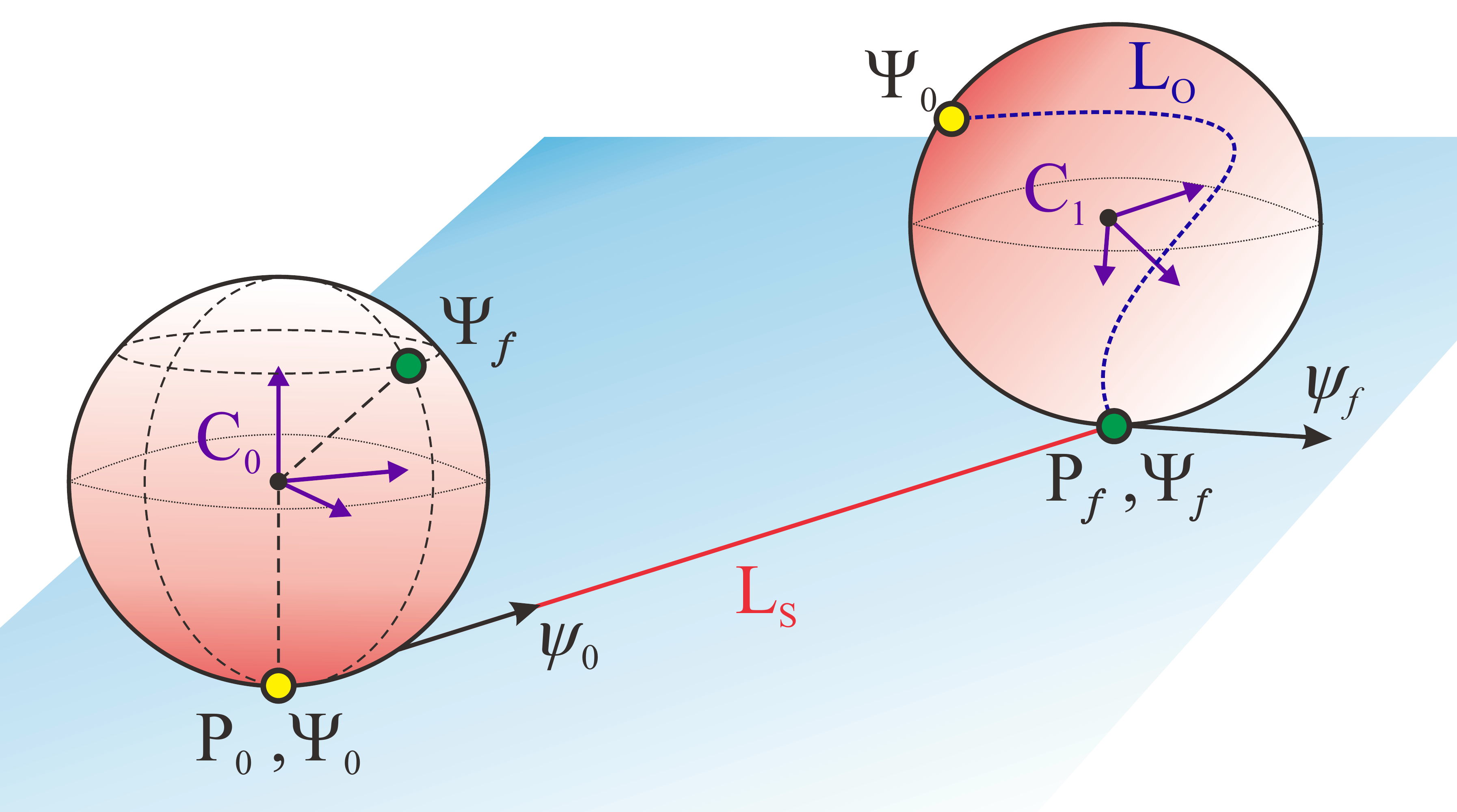}		
	\caption{Statement of planning problem while the sphere with contact path of $\uvec{L}_{o}$ follows a straight path $\uvec{L}_{s}$ on plane.}\label{Fig:statementPlane}
\end{figure}
Configuration of a sphere on a plane (see Fig. \ref{Fig:statementPlane}) is described with the sphere position on plane $\uvec{P}$=$(u_{s},v_{s})$ and its local orientation $\bm{\Psi}=(u_{o},v_{o})$ which $\psi$ is the spin angle between the sphere and plane. We consider our initial and final states with following notations $\{\uvec{P}_0$, $\bm{\Psi}_0$, $\psi_0\}$$=\{u_{s,0}$, $v_{s,0}$, $u_{o,0} $, $v_{o,0} $, $\psi_0 \}$ and  $\{\uvec{P}_f$, $\bm{\Psi}_f$, $\psi_f\}$$=\{u_{s,f}$, $v_{s,f}$, $u_{o,f}$, $v_{o,f}$, $\psi_f \}$, respectively. The coordinate system and preliminary information of this ball-plate system is described in \ref{Basicballplate}. In this planning, the state equation is solved in $k$ iterations till finding admissible paths, where the traveled paths on the sphere and plane are noted as
$\uvec{L}_{o}$ and $\uvec{L}_{s}$. We plan the sphere to reach its final configuration $C_0 \rightarrow C_1$ with a straight trajectory on $\uvec{L}_s$ in the given $t_f$ time.

It is assumed that the initial states $\{\uvec{P}_0,\bm{\Psi}_0,\psi_0\}$ and the final desired states of the sphere $\{\bm{\Psi}_f,\psi_f\}$ are given. However, the final desired position on the plane $\uvec{P}_f$ has to be chosen under the consideration that the length between initial and final position $||\uvec{P}_{f}-\uvec{P}_{0}||_2=[(u_{s,f}-u_{s,0})^2+(v_{s,f}-v_{s,0})^2]^{\frac{1}{2}}$ is greater than a minimum distance variable $d$ where this constraint is calculated by using the rest of desired and initial states. This length limitation appears because the total curve length of $\uvec{L}_{o}$ on the sphere is always the same (no-sliding constraint) as $\uvec{L}_{s}$ that goes in a straight trajectory. Also, the sphere has to arrive at desired local coordinate $\bm{\Psi}_f$ with different approaching angles, as the desired spin $\psi_f$, while it follows a straight optimal line rather than maneuvering freely through the plane $U_S$. Hence, by knowing the sphere desired states $\{\bm{\Psi}_f,\psi_f\}$, plane's desired states $\uvec{P}_f$ are chosen in any desired direction $\{u_{s,f},v_{s,f}\}$ with the length $||\uvec{L}_s||=||\uvec{P}_{f}-\uvec{P}_{0}||_2$ larger than the minimum distance variable $d$ as
\begin{equation}
d<||\uvec{P}_{f}-\uvec{P}_{0}||_2,
\label{Eq:minimumditancerule}
\end{equation}
Note that this constraint 
is not important and the sphere can reach all possible configurations, if the length of the desired position $||\uvec{P}_{f}-\uvec{P}_{0}||_2$ is set larger than sphere circumference $2\pi R_o$.

In order to find the minimum distance variable $d$, a path is constructed as a circular segment on the sphere that passes $\bm\Psi_f$, and then the cap area is changed toward true desired $\psi_f$ by Gauss-Bonnet theorem \cite{DIfgeometry1976}. Then, by the trigonometric relations in the new under-cap area, we find the length of $\uvec{L}_o$ as the minimum distance,
where $||\uvec{L}_s||=||\uvec{L}_o||$.
\begin{figure}[t!]
	\centering
	\includegraphics[width=3.5 in]{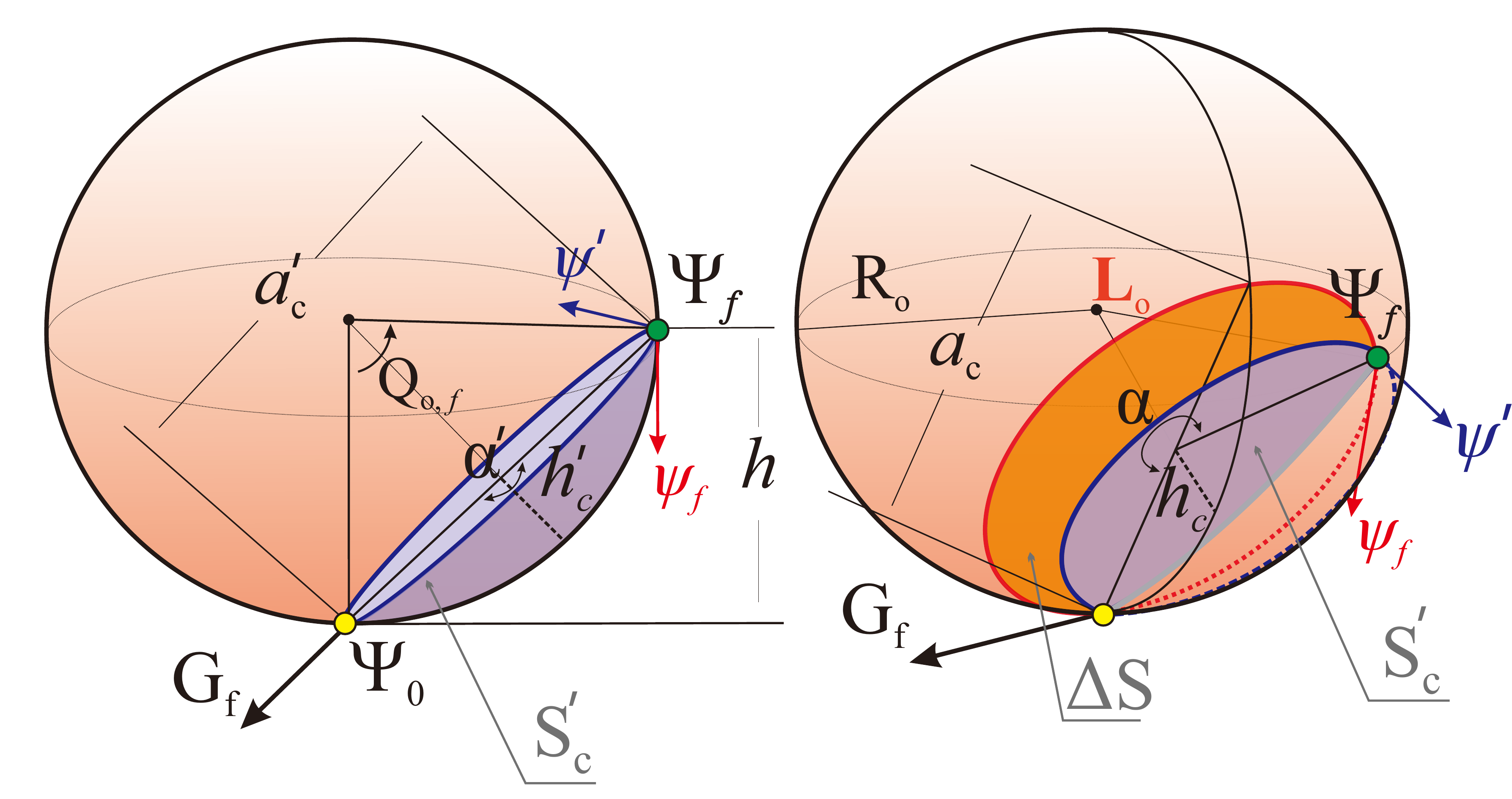}		
	\caption{Limit of minimum distance $d$ that is required to reach desired $\psi_f$. Note that $G_f$ shows the desired goal angle on plane. }\label{Fig:SphereDistanceMotion2}
\end{figure}

Let the minimum distance $d = 2\pi a_c \alpha$ be the circumference of the orange cap-area that sphere travels, red line $\uvec{L}_o$ on the right sphere at Fig. \ref{Fig:SphereDistanceMotion2}, where $a_c$ and $\alpha$ are the base diameter of this circumference and the angle of sector from $\bm\Psi_{0}$ to $\bm\Psi_f$ on the sphere $U_C$, respectively. First, we find the base diameter $a_c$ of the cap with the following formulation
\begin{align}
a_c=2\sqrt{(S_t/\pi)-h^2_c},\;\;\;h_c=S_t/(2\pi R_{o}),
\label{Eq:Finalcapparameters}
\end{align}
where $S_t$ and $h_c$ are the total area of the cap-shaped region [orange and blue regions in Fig. \ref{Fig:SphereDistanceMotion2}] and height of the cap area. Notice that the rotation of the sphere along the red path $\uvec{L}_o$ doesn't cover the whole cap-shaped region. Thus, we find the total area as $S_t=2S'_c+\Delta S$ where $S'_c$ is the constructed cap area of the circular path by $\bm\Psi_f$ [see the blue cap-shaped region in Fig. \ref{Fig:SphereDistanceMotion2}] and $\Delta S$ is the area change [see the orange region in Fig. \ref{Fig:SphereDistanceMotion2}] for reaching from the spin angle $\psi'$ with cap area of $S'_c$ to desired spin angle $\psi_f$. Here, $S_t$ has two parts which combining them gives the full segment of the circular cap (dashed and solid red line $\uvec{L}_o$). First, the area of the cap-shaped region of $S'_c$ is determined from the closed simple circle that passes $\bm\Psi_f$
\begin{align}
S'_c=\left(\alpha'/2\pi\right)   \left[ ({a}'_c/2)^2+{h'}_c^2 \right],
\label{Eq:AreaCAp}
\end{align}
where ${a'}_c$, ${h'}_c$ and $\alpha'$ are the diameter of cap's base, height of the cap and the sector angle from sphere initial to final configuration $\bm{\Psi}_f$ on the cap's base, $\alpha'$ equals to $\pi$. Following parameters in (\ref{Eq:AreaCAp}) are calculated with the help of Eq. (\ref{Eq:Coorinatequationsphereonplane}) and trigonometric relations, shown in Fig. \ref{Fig:SphereDistanceMotion2},
\begin{equation*}
\begin{split}
&h=R_{o}\left[1-\cos u_{o,f} \cos v_{o,f} \right],\;	{a'}_c=\left[h^2+R_o^2\left(\sin^2 v_{o,f} + \sin^2 u_{o,f} \cos^2 v_{o,f} \right)\right]^{1/2},\;\\
&{h'}_c=\begin{cases}
&R_{o}\left[1-\cos \left(Q_{o,f}/2 \right)\right],\;\;\;\;\;\;\;\;\;\;\;\;\;\;\;\;\;\;{a}'_c \leq 2R_{o}\\
&R_{o}\left[ 1- {a}'_c \cos \left(Q_{o,f}/2 \right)/2R_{o}\right],\;\;\;\;\;\;  {a}'_c > 2R_{o}
\end{cases}
\end{split}
\end{equation*}
where $Q_{o,f}$ is the angle from initial contact point $\bm{\Psi}_{0}=\{0,0\}$ to desired local coordinates $\bm{\Psi}_{f}$ as
\begin{equation*}
Q_{o,f}=\begin{cases}
&\pi-2\cos^{-1}\left(h/a'_c\right),\;\;\;\;\;\;\;\;\;\;\;\;\;\;\;\;\;h \leq  R_{o}\\
& \frac{\pi}{2}-\sin^{-1}\left((h-R_o)/R_o\right),\;\;\;\;\;\;\;\;\hfill h >  R_{o}
\end{cases}
\end{equation*}
Also, by knowing the $a_c$ and $a'_c$, the new sector angle from $\bm\Psi_{0}$ to $\{\bm\Psi_f,\psi_f\}$ ( $\uvec{L}_o$ as the red line in Fig. \ref{Fig:SphereDistanceMotion2}) becomes
\begin{equation}
\alpha=\frac{1}{2\pi} \cdot
\begin{cases}
& 1- 2 \sin^{-1}(a'_c/a_c), \;\;\;\;\;\;\;\;\;\;\;\Delta S \geq 0\\ 
&  2 \sin^{-1}(a'_c/a_c),	 \;\;\;\;\;\;\;\;\;\;\;\;\;\;\;\;\;\Delta S < 0
\end{cases}
\label{EQ:Lfexludedone}
\end{equation}
To find the achieved spin angle $\psi'$ by the contacted $S'_c$ cap-area, we use the Gauss-Bonnet theorem \cite{DIfgeometry1976} while the sphere touches the circular closed blue path
\begin{equation}
\Delta \psi= \psi'-\psi_0=\iint_{{S}'_{c}} \kappa_{o} \; dS = S'_c/R_{o}^2,
\label{Eq:Gauss-Bonnetbase}
\end{equation}
\begin{figure}[t!]
	\centering
	\includegraphics[width=3.7 in]{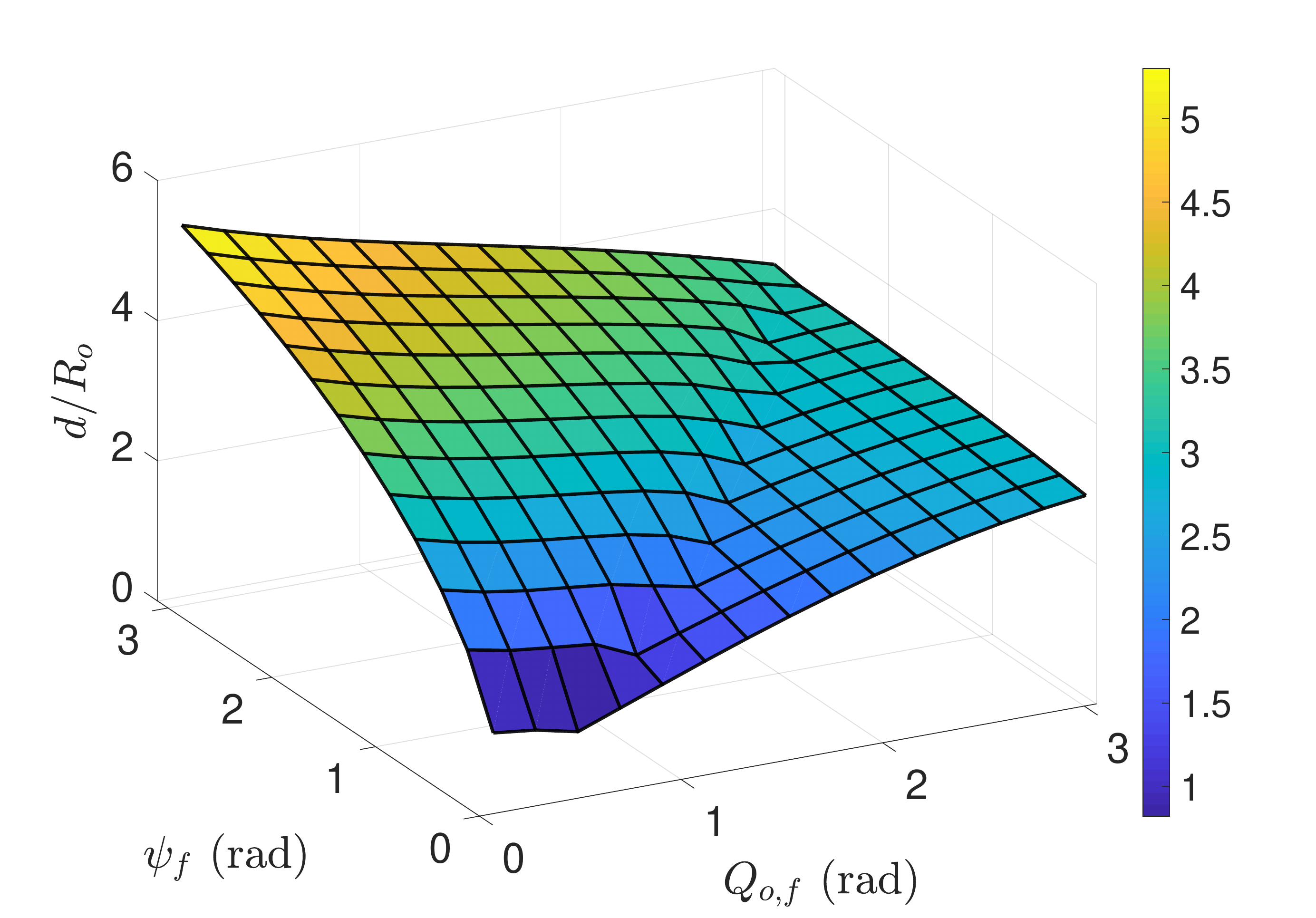}		
	\caption{Normalized minimum distance $d/R_o$ for different desired angles where $u_{o,f}\in[0,\pi]$ and $v_{o,f}= 0.01$.}\label{Fig:DISRATIO1}
\end{figure}where $\kappa_{o} =1/R^2_{o}$ is the Gaussian curvature and the initial spin angle is assumed $\psi_0=0$. Finding the spin angle $\psi'$ also let us calculate $\Delta S$ in $S_t$. The required area-change $\Delta S$ for achieving $\psi_f$ is calculated by the same Gauss-Bonnet theorem as follows
\begin{equation}
\Delta S = (\psi_f-{\psi'})/\kappa_{o} = R^2_{o}(\psi_f-{\psi'}).
\end{equation}
After obtaining the total area $S_t$ for reaching $\psi_f$, $a_c$ in (\ref{Eq:Finalcapparameters}) and $\alpha$ in (\ref{EQ:Lfexludedone}) give us the value of minimum distance $d$. Fig. \ref{Fig:DISRATIO1} depicts the normalized minimum distance for different desired angles which shows how the small distance $||\uvec{P}_{f}-\uvec{P}_{0}||_2$ can be for the example final configurations. It is clear that as the desired local coordinate $\bm\Psi_f$ moves to upper-hemisphere, $\uvec{P}_f$ has to set for the longer length from $\uvec{P}_0$. Also. the same property is true for rising the spin angle $\psi_f$. Moreover, the largest distance requirement happens at lower points of $\bm\Psi_{f}$ ($Q_{o,f} \leq \pi/2$) with larger desired spin angles $\psi_f$. Note that this constraint (\ref{Eq:minimumditancerule}), with the assumption of a simple circular curve, is looked as the shortest length under the isoperimetric inequality, $4 \pi S_t \leq d^2$.

\section{Geometric Controller for Darboux-Frame-Based Kinematics}
\label{Sub:sectionVirtual}
We utilize the model of the Darboux frame at the contact point of the spin-rolling sphere and plane presented in  \cite{Tafrishi2021DarKiF}. The kinematics readily suits both contact trajectories and arbitrary parameters of the surfaces \cite{cui_sun_dai_2017,CuiDarboux2010,Riemannian2002}. Also, this transformation provides two significant benefits through our planning approach. First, the spin-rolling angular rotations explicitly appear on the relative curvature and torsion \cite{CuiDarboux2010} that makes it easier for manipulation.  Second, the Darboux frame separates the time variable from the planning due to its time- and coordinate-invariance.

As show in \cite{Tafrishi2021DarKiF}, the transformed fully-actuated Darboux-frame-based kinematics model of a spin-rolling sphere on a plane in the time domain is presented as
\begin{align}
\begin{split}
&\left[\begin{array}{c}
\dot{u}_{s}(t)\\
\dot{v}_{s}(t)\\
\dot{u}_{o}(t)\\
\dot{v}_{o}(t)\\
\dot{\psi}(t)
\end{array}\right]=\delta \Bigg ( \left[\begin{array}{c}
\sin(\theta+\varphi)\\
\sin(\theta+\varphi)\\
\frac{\sin(\theta+\varphi)[\sin{\psi}-\cos{\psi}]}{R_{o}\cos{v_{o}}}\\
\frac{\sin(\theta+\varphi)[\cos{\psi}+\sin{\psi}]}{R_{o}}\\
\frac{\tan{{v}_{o}}[\sin(\theta+\varphi)(\sin\psi-\cos\psi)+\cos\varphi]}{R_{o}}
\end{array}\right]+
\left[\begin{array}{c}
-R_{o}\sin(\theta+\varphi)\\
-R_{o}\sin(\theta+\varphi)\\
\frac{\sin(\theta+\varphi)[\cos{\psi}-\sin{\psi}]}{\cos{v_{o}}}\\
-\sin(\theta+\varphi)[\sin{\psi}+\cos{\psi}]\\
\tan{{v}_{o}}[\sin(\theta+\varphi)(\cos\psi-\sin\psi)]
\end{array}\right]\;\gamma_{s}\\
&
+
\left[\begin{array}{c}
R_{o}\sin(\theta+\varphi)\\
-R_{o}\cos(\theta+\varphi)\\
\frac{-\sin{(\psi+\theta+\varphi)}}{\cos{v_{o}}}\\
-\cos{(\psi+\theta+\varphi)}\\
-\tan{{v}_{o}}\sin{(\psi+\theta+\varphi)}
\end{array}\right]\;\beta_{s}+\left[\begin{array}{c}
0\\
0\\
0\\
0\\
-1
\end{array}\right]\;\alpha_s \Bigg ).
\end{split}
\label{EQ:LatestStateEquation}
\end{align}
where five state system $\{\uvec{P}(t),\bm{\Psi}(t),\psi(t) \}$ has $\{\alpha_s,\beta_s,\gamma_s,G_f, \delta\}$ inputs which
\begin{align}
&\theta(\beta_s,\gamma_s,G_f)=\cot^{-1} \Big [\frac{1}{\beta_{s}}\Big(
\frac{1}{R_{o}}(1-\tan G_f) +\gamma_{s}(-1+\tan
G_f)-\beta_{s}\tan{G_f} \Big) \Big]-\psi_q, \nonumber\\
&  \varphi(G_f)=\psi_q+\begin{cases}
&\pi, \;\;\;\;\;\;\;\;\;\;\;-\frac{3\pi}{4}<G_f<0 \;\;\;\;\&\;\;\;\;\;\; 0 \leq G_f<\frac{\pi}{4}\\
&0,\;\;\;\;\;\;\;\;\;\;\;\;-\pi<G_f<-\frac{3\pi}{4} \;\& \;-\pi \leq G_f<\frac{\pi}{4}
\end{cases}.
\label{Eq:AngleGoalthetaM}
\end{align}
where $\psi_q$ is the spin angle deviation. The three arc-length-based control inputs $\{\alpha_s,\beta_s,\gamma_s\}$
correspond to the angular velocities of the sphere in time domain. The remaining angular $G_f$ and rolling rate $\delta$ inputs are for directing sphere on fixed surface (plane) while they are constraining virtual surface inputs $\{\alpha_s,\beta_s,\gamma_s\}$. The controllability of this geometric model (\ref{EQ:LatestStateEquation}) is in \cite{Tafrishi2021DarKiF}. 
This geometric-based controller converges the rotating object (sphere) trajectory to the desired angular states of the sphere while $G_f$ keeps the sphere on suitable direction with rolling rate $\delta$. In here, we define the goal angle by $G_f$=$\tan^{-1}\big [(v_{s,f}-v_{s,0})$$/(u_{s,f}-u_{s,0})\big ]$; hence, $\theta$ and $\varphi$ will change relative to any values of arc-length inputs, for keeping the sphere always along $G_f$ angle to reach $\uvec{P}_f$. Therefore, the sphere stays in the prescribed straight direction by substituting constant value to $G_f$ from the desired position on the plane $\uvec{P}_f$.

The Darboux-frame-based kinematics (\ref{EQ:LatestStateEquation})-(\ref{Eq:AngleGoalthetaM}) with the inputs in the arc-length domain requires a geometric control for converging spin-rolling sphere to desired angular configuration $\bm\Psi_f$. We introduce a virtual surface $U_V$ to define these arc-length-based inputs $\{\alpha_s,\beta_{s},\gamma_{s} \}$. The virtual surface is a surface sandwiched between sphere and plane at the contact frame. From a physical point of view, deformation (curvature changes) of this virtual surface is projected onto both sphere and plane trajectories \cite{Tafrishi2021DarKiF}. These changes in the curvature manipulates the curve on the sphere $\uvec{L}_o$ like a flexible rope. To bend this rope-like curve toward the desired full-configuration, we propose a tuning algorithm to update the variables of this virtual surface in the incoming section.

\begin{figure}[t!]
	\centering
a)	\includegraphics[width=2.9 in,height= 1.7 in]{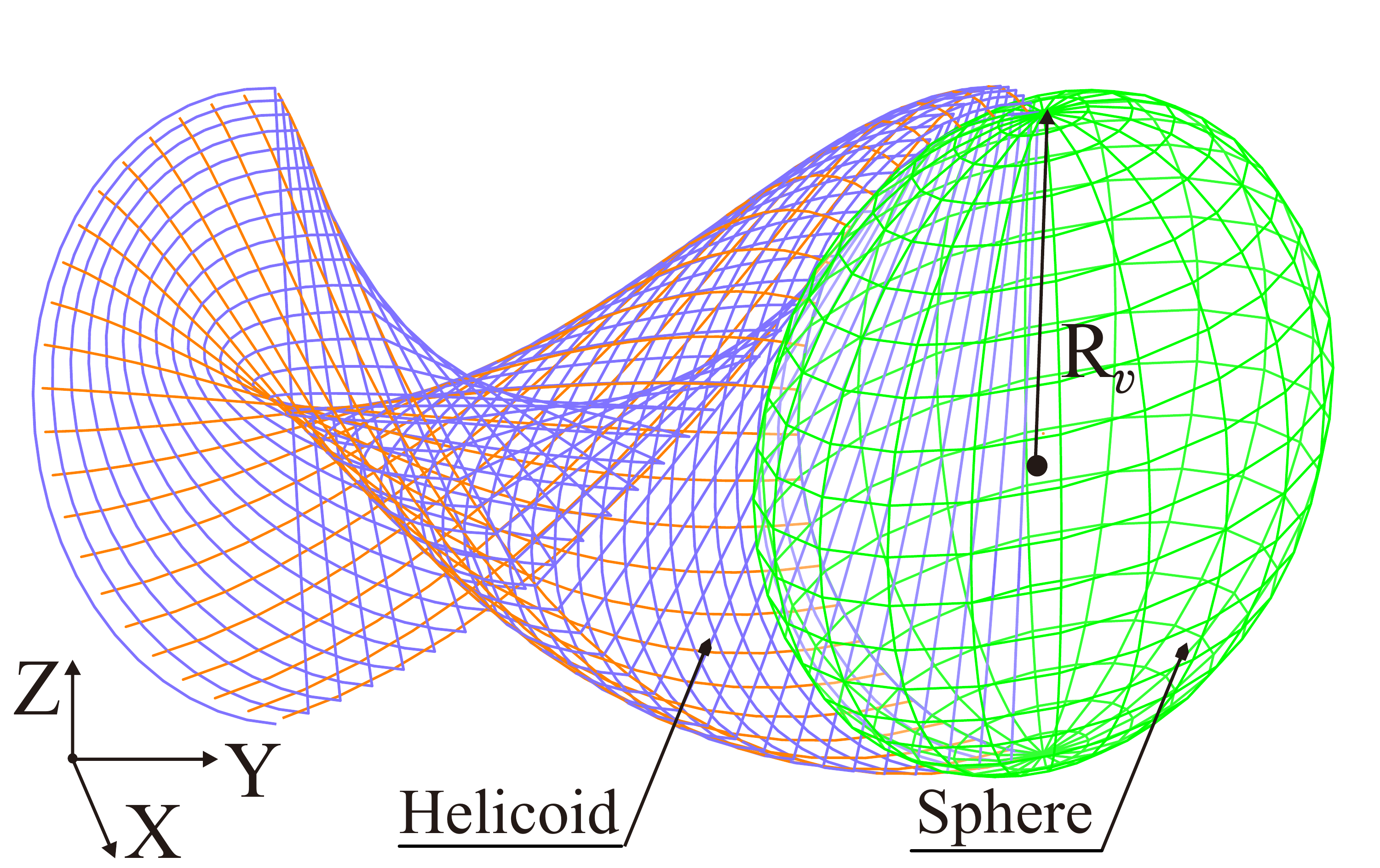}	
b)	\includegraphics[width=2.8 in]{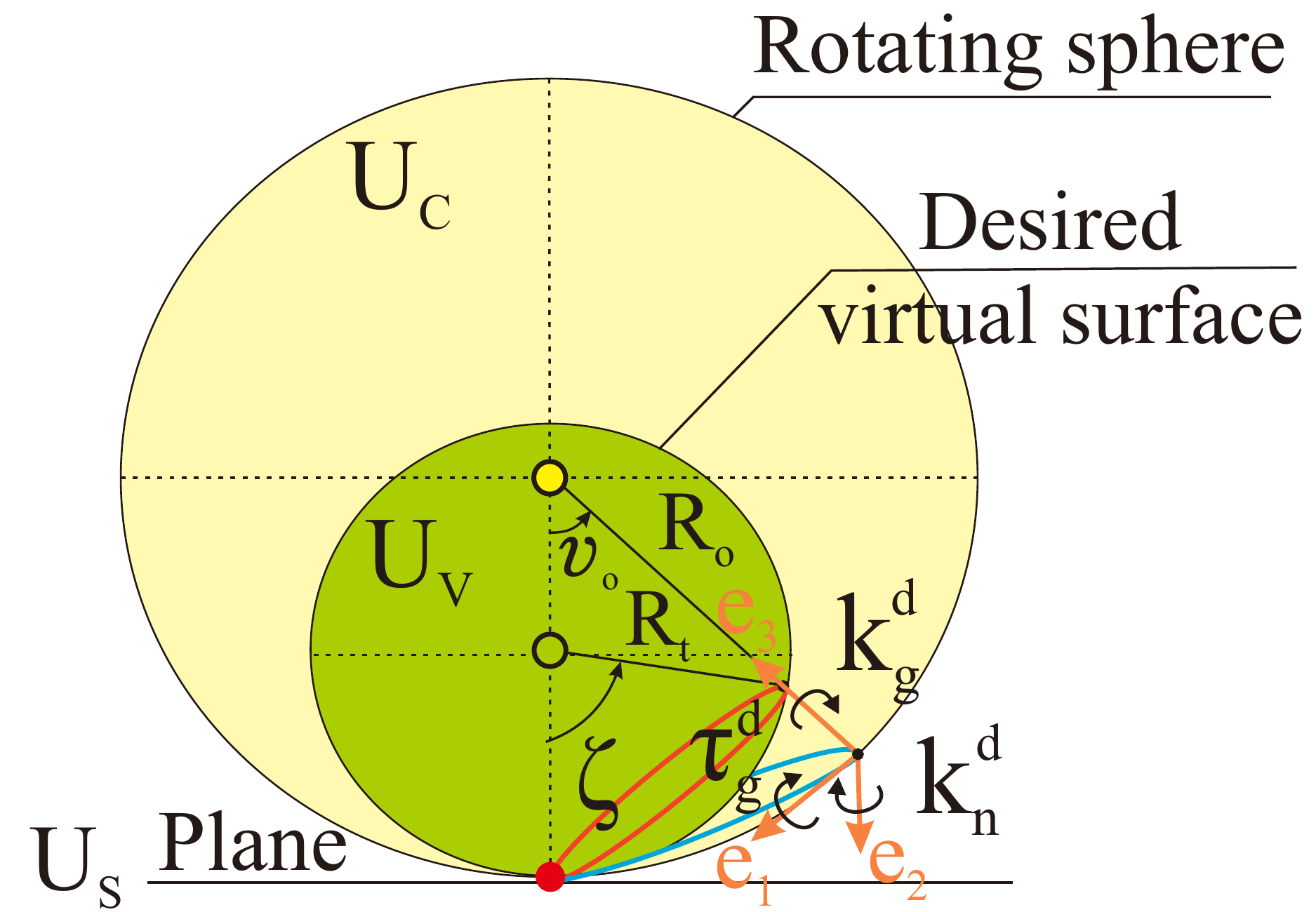}	
	\caption{a) Geometric shape of the used virtual Surface $U_V$. Note: Green and blue surfaces are for $R_t=0$ as the sphere and $R_t=R_{v}$ as the helicoid, b) Design of the arc-length-based inputs by using desired virtual surface. Note that schematic is drawn for $u_{o}=0$.}\label{Fig:Virtualsurface123}\label{Fig:RCCO}
\end{figure}
Here, we design the virtual surface for controlling the curve $\uvec{L}_{o}$ on the sphere $U_C$ when $\uvec{L}_s$ is specified as a straight path.
	In order to develop a geometric controller based on the virtual surface, we use the induced curvature (differential characteristics), with the geodesic curvature $k_g$, the normal curvature $k_n$ and the geodesic torsion $\tau_g$, that consists of the rolling convex object (sphere) on a plane with sandwiched virtual surface as follows \cite{Tafrishi2021DarKiF}
    \begin{eqnarray}
	k_g= k^{o}_g-k^{s}_g-\alpha_s,\;\tau_g=\tau_g^{o}-\tau^{s}_g-\beta_{s}, \;k_n=k^{o}_n-k^{s}_n-\gamma_{s},
	\label{EQ:TheCurvaturerelevantDif}
	\end{eqnarray}
	where $\{k^{o}_g,\tau^{o}_g,k^{o}_n\}$, $\{k^{s}_g,\tau^{s}_g,k^{s}_n\}$ and $\{\alpha_s,\beta_s,\gamma_s\}$ are the geodesic curvature, geodesic torsion and normal curvature for, respectively, the rolling object (sphere), plane and virtual surface. In the geometric controller, the curvature properties are designed with similarity to rotating object (sphere) manifold $U_C$  (\ref{Eq:detailMontanaCOonplane}) for planning on the virtual surface. However, the moving object is a sphere with $\tau^{o}_g=0$. This cause the model (\ref{EQ:LatestStateEquation}) to be unstable \cite{Tafrishi2021DarKiF} because $\beta_{s}=0$. As a solution, geodesic torsion of a helicoid shape is applied where the geometric shape is shown in Fig. \ref{Fig:Virtualsurface123}-a (see \ref{VirtualSurfaceGeodesicTorsionProof} for details of derivation)
	\begin{equation}
	\tau^{v}_g=\frac{1}{R_{v}^2}|R_{v}^2\cos^2{v_{v}(t)}-R^2_t|^{\frac{1}{2}},
	\label{Eq:GEODESICTorsionVirtual}
	\end{equation}
	where $R_{v}$, $R_t$ and $v_v(t)$ are defined by main spherical radius, sum of spherical and torsion radii and $v$-curve angle of helicoid surface, respectively. As a keynote, geometric surface shown in Fig. \ref{Fig:Virtualsurface123}-a is able to transform from the sphere to the helicoid surface by changing $R_t$.

	Next, we utilize (\ref{EQ:TheCurvaturerelevantDif}) to construct the desired virtual surface (see Fig. \ref{Fig:RCCO}-b) with the normal and geodesic curvature of the spherical surface $\{k_n^d(s),k_g^d(s)\}$ in (\ref{Eq:detailMontanaCOonplane})-(\ref{Eq:MontanaCSNN}) and the geodesic torsion of the helicoid surface $\tau^d_g(s)$ (\ref{Eq:GEODESICTorsionVirtual}) as follows
	\begin{eqnarray}
	k_g^d(s)= \tan{\zeta}/R_t=\tan{v_{o,f}}/R_{o}-\alpha_s,\;
	\tau^d_g(s)=\frac{1}{R^2_{o}}\left|R^2_{o}\cos^2{v'_{o}}-R^2_t\right|^{\frac{1}{2}}=\beta_{s},\;	 k_n^d(s)=1/R_{n}=1/R_{o}-\gamma_{s}.
	\label{Eq:TheCurvaturepropertymodeldesign}
	\end{eqnarray}
	where $R_t=R_{n}+R_{g}$ and $v'_o= v_{o,f}-v_o(t) $ are the total radius and the angle feed of $v$-curve in which $R_n$ and $R_g$ are the desired normal curvature and geodesic torsion radii. Also, we assume $\zeta$ is the desired stereographic projection angle for the spherical object along $k^d_g \bm{e}_3$ vector. By considering that the diameter of projected curve is same (red and blue curves in Fig. \ref{Fig:RCCO}-b ) and $R_t\tan\zeta=R_o\tan v_o $, $\zeta$ is determined
	\begin{equation}
	\zeta=\tan^{-1}\left[R_{o}\tan \left( v_{o,f}+\zeta' \right)/R_t\right],
	\end{equation}
	where $\zeta'$ is the constant angle shift that will be used during planning updates. Also, $R_n$ and $R_g$ are defined as
	\begin{align}
	R_n(t)=R_g(t)=\left[ R_i(t)+R_a \right]/2,
	\end{align}
	\begin{figure}[t!]
		\centering	
		\includegraphics[width=3 in]{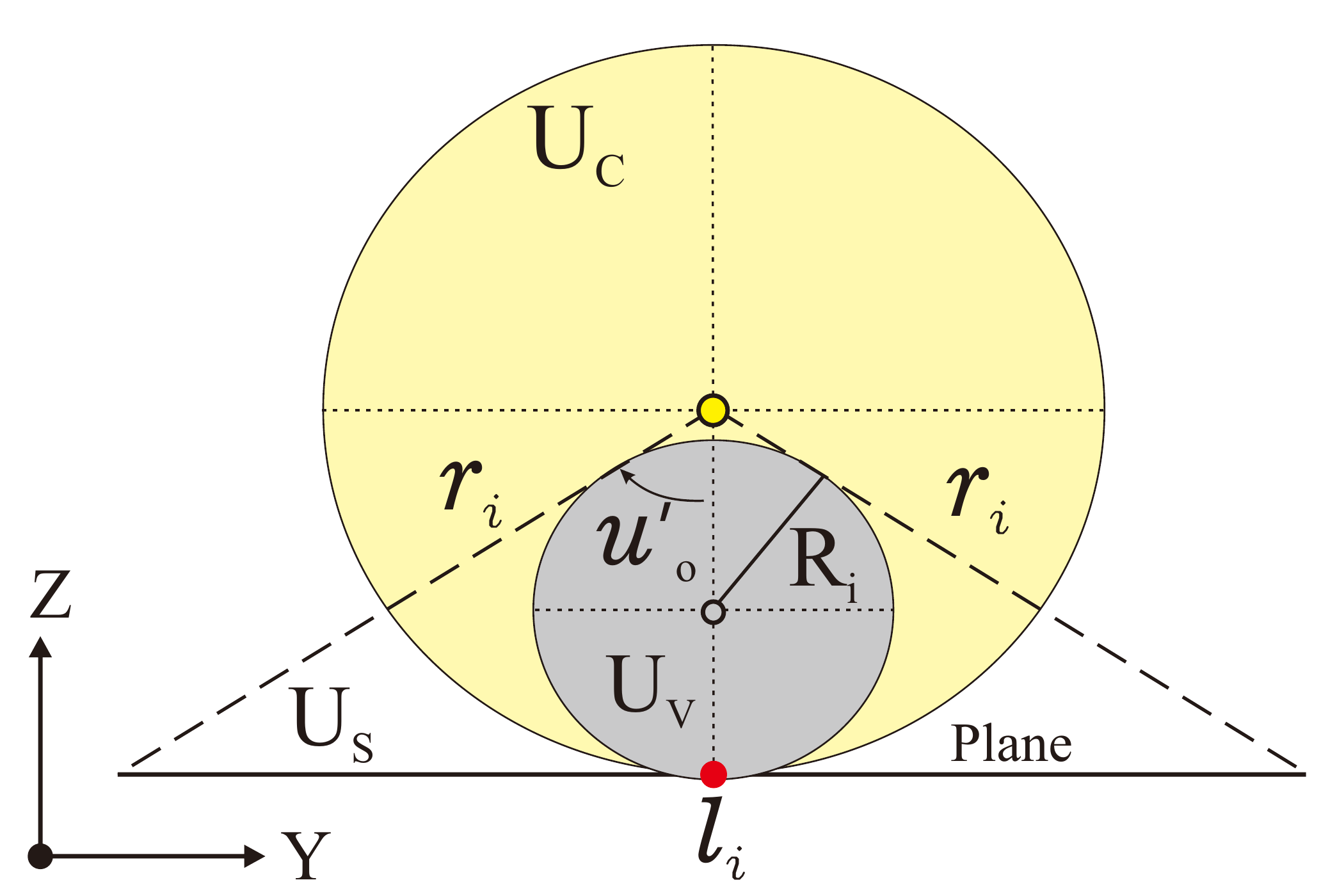}				
		\caption{$R_i$ determination by incircle of $u'_{o}$.}\label{Fig:QCOHeeightttt}
	\end{figure}
where $R_i$ and $R_a$ are the built-in circle radius of $u$-curve and tuning constants that happens by algorithm, respectively. By using schematic of Fig. \ref{Fig:QCOHeeightttt}, built-in radius $R_i$ is computed with the change of the designed incircle radius as follows
	\begin{equation}
	R_i(t)=
	\begin{cases}
	\begin{split}
	&\left[\frac{(S_i-r_i)^2(S_i-l_i)}{S_i}\right]^{\frac{1}{2}},\;\;\;\;\;\;\;\;\;\;\;0 \leq u'_{o}<\frac{\pi}{2}\\
	&\frac{R_o}{\mu_r}+\left[\frac{(S_i-r_i)^2(S_i-l_i)}{S_i}\right]^{\frac{1}{2}},\;\;\frac{\pi}{2}\leq u'_{o} \leq \pi
	\end{split}
	\end{cases}
	\label{EQ:HeightQCO}
	\end{equation}
where $S_i=(2r_i+l_i)/2$ is the area of encompassed triangle of the incricle, $u'_{o}=u_{o,f}-u_o(t)$ is the convergence of $u$-curve angle, $\mu_r$ is the scaler to limit the maximum built-in radius $R_i$, and also $r_i=R_{o}/\cos u'_{o}$ and $l_i=2R_{o}\tan{u'_{o}}$ are adjacent and hypotenuse sides of isosceles triangle (see Fig. \ref{Fig:QCOHeeightttt}).
	\label{DefintionofDesireVirtualSurface}
\begin{rem}
	The sphere curvature in (\ref{Eq:TheCurvaturepropertymodeldesign}) is included to cancel out the existing sphere $U_C$ properties that drift terms is presenting in the kinematic model (\ref{EQ:LatestStateEquation}). Thus, this design gives a direct manipulation of virtual surface curvatures (s-domain) on $\uvec{L}_o$ trajectory for $\bm{\Psi}_f$.
\end{rem}
Now, we can find the geometric control inputs from (\ref{Eq:TheCurvaturepropertymodeldesign}) as
\begin{eqnarray}
\alpha_s=\tan v_{o,f}/R_{o}-\tan\zeta/R_t,\;\beta_{s}=\frac{1}{R^2_{o}}|R^2_{o}\cos^2{v'_{o}}-R^2_t|^{\frac{1}{2}},\;\gamma_{s}=\frac{R_n-R_{o}}{R_nR_{o}}.
\label{EQ:VirtualSurfaceGeodesicinputmain}
\end{eqnarray}
By substituting the desired values $(u_{o,f},v_{o,f})$, the curves of $\uvec{L}_o$ are constrained on the virtual surface with the projection on the rotating object, as the example of Fig. \ref{Fig:RCCO}. This constrained curve is created from a conservative vector field in (\ref{EQ:VirtualSurfaceGeodesicinputmain}) where independent inputs act as force fields $\uvec{F}= \nabla \uvec{E}_{in}(u_o(t),v_o(t),u_{o,f},v_{o,f})=\{ \alpha_s, \beta_{s}, \gamma_s \}$ to manipulate at contact frame on $U_C$ manifold \cite{tapp2016differential}. By using kinematics (\ref{EQ:LatestStateEquation})-(\ref{Eq:AngleGoalthetaM}), it is interpreted that the desired virtual surface bends this vector field towards the desired local coordinate $\bm{\Psi}_f$.
Note that our curve manipulation on the sphere manifold $U_C$ can be imagined as a flexible rope with length of $||\uvec{P}_f-\uvec{P}_0||_2$ that $\{R_n,R_g\}$ and $\zeta$ terms change the curvature radius and angle of $v$-curve for  $\uvec{L}_o$ trajectory on the spherical surface $U_C$.

It should also be noted that the proposed virtual surface is designed in a way that eases our planning problem with spherical curvature properties $k^{v}_n$ and $k^{v}_g$. However, more complicated virtual surfaces with existing geodesic torsion $\tau^{v}_g$ can be applied to this planning approach for creating different contact paths but it can complicate planning problem,  e.g., algorithm tuning, which requires a separate study.

\section{Motion Planning Algorithm}
\subsection{Iterative Tuning Algorithm}	
\begin{figure}[t!]
	\centering	
a)	\includegraphics[width=3.2 in]{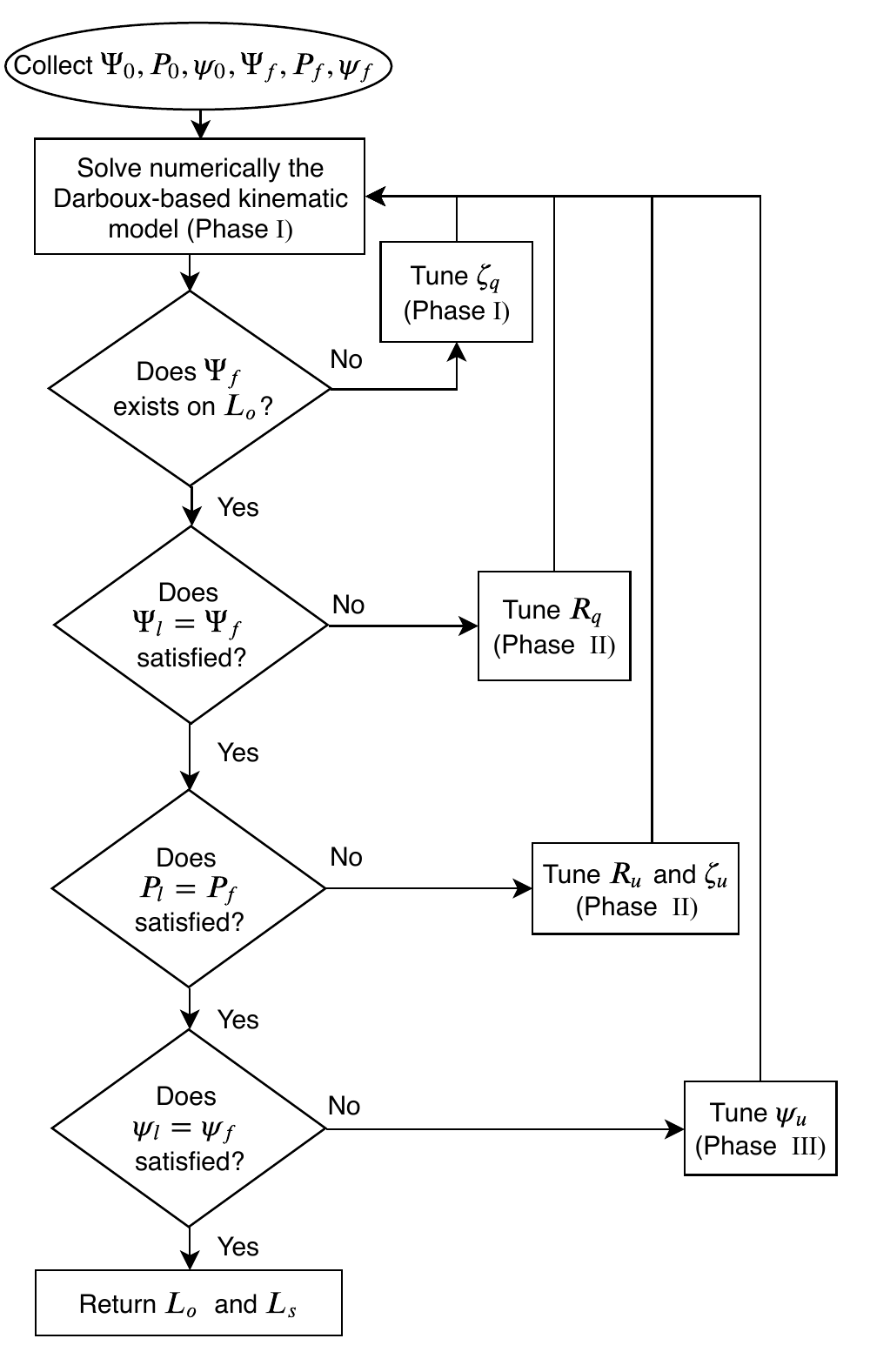}	
b)	\includegraphics[width=2.8 in,height= 1.9 in]{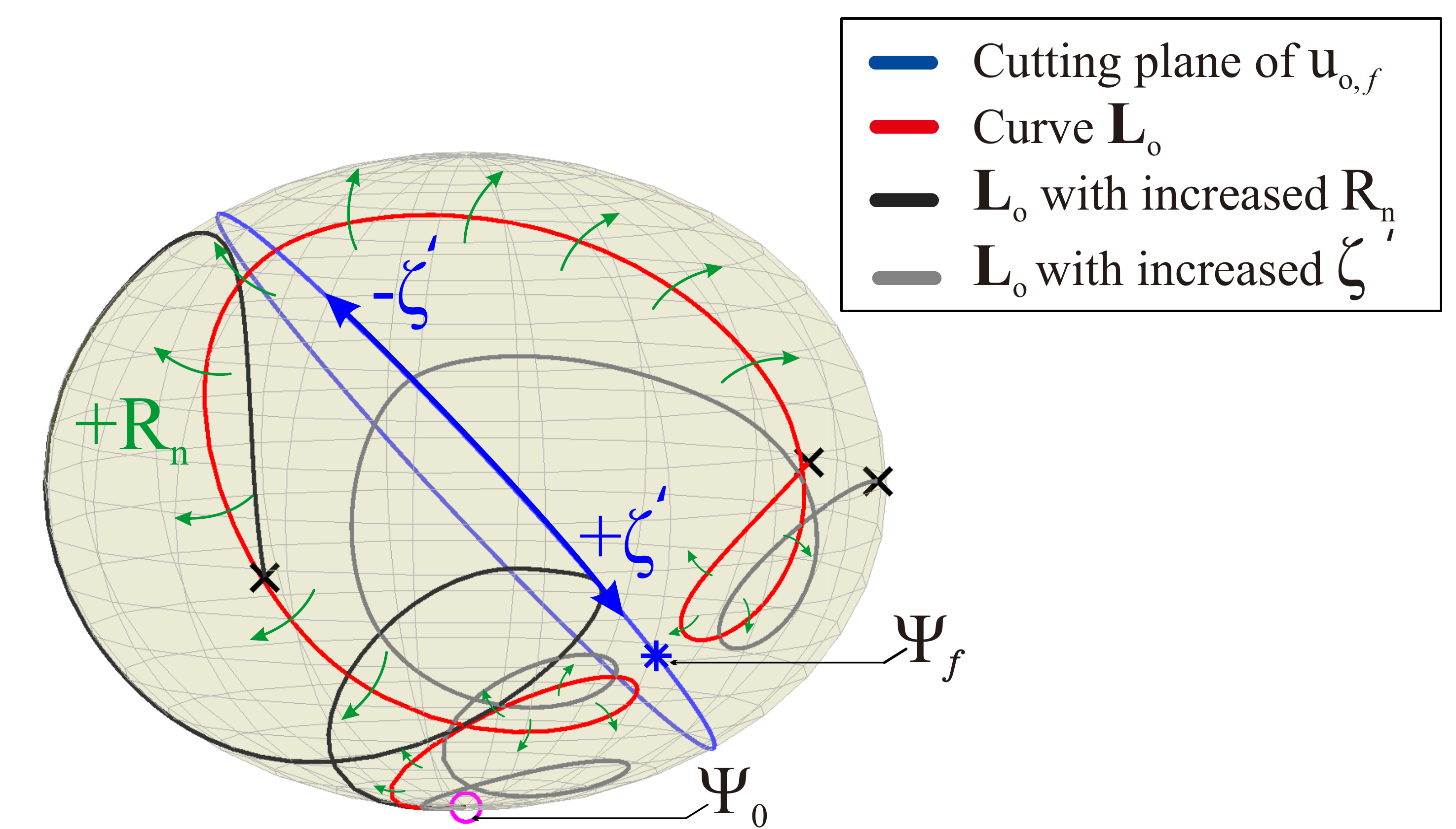}			
	\caption{a) The flowchart of planning algorithm, b) Manipulation of the rope-like $\uvec{L}_o$ curve  with the designed virtual surface.Note that following curves are obtained by using formulas in Phase I for $\psi_q $ and $\delta$ in Eq. (\ref{Eq:PureSpinAngleDEviation}) and (\ref{Eq:DELTAEQuationtotal}).}\label{Fig:BLOCKDIAGRAMFeedForward}
	\label{Fig:Vector_field3}
\end{figure}
We plan the motion of the sphere spinning on the plane along the straight line $\uvec{L}_{s}$ with reaching a desired orientation. The flowchart in Fig. \ref{Fig:BLOCKDIAGRAMFeedForward}-a shows our algorithm that uses the control inputs (\ref{EQ:VirtualSurfaceGeodesicinputmain}). Geometrically, it can be considered as manipulating curvature radii $\{R_n,R_g\}$ and angular location $\zeta'$ to bend the created rope-like curve $\uvec{L}_o$ from (\ref{EQ:VirtualSurfaceGeodesicinputmain}) on the sphere. For example, constant raise in $\{R_n,R_g\}$ expand the curve $\uvec{L}_o$ like enlarging the loops of the rope while its length is same [see Fig. \ref{Fig:Vector_field3}-b]. Also, constant changes on $\zeta'$ moves the curve $\uvec{L}_o$ on the cutting plane of $u_{o,f}$. To do the convergence of this rope-like model, the proposed algorithm will solve (\ref{EQ:LatestStateEquation})-(\ref{Eq:AngleGoalthetaM}) and (\ref{EQ:VirtualSurfaceGeodesicinputmain}) in iterations with the re-tuned constants in $\{R_n,R_g\}$ and $\zeta'$.

In general, to shift the curve of the sphere $\uvec{L}_o$ by $\zeta'(k)=\zeta_q (k) +\zeta_u (k)$ on the cutting plane $u_{o,f}$, updates happen in Phase I and II by $\zeta_q$ and $\zeta_u$, respectively. Also, constant change in the desired radius $R_a$ is defined by the inclusion of all phases as
$R_a(k)= R_q(k)+R_u(k)$, where $R_q$ and $R_u$ are tuning constants of radii in the steps of the Phase II. Our tuning algorithm consists of three primary steps:

The first step, Phase I, is the main part of the algorithm where the kinematic model (\ref{EQ:LatestStateEquation})-(\ref{Eq:AngleGoalthetaM}) with designed arc-length-based inputs (\ref{EQ:VirtualSurfaceGeodesicinputmain}) is numerically solved in the time domain. After obtaining $\uvec{L}_o$ and $\uvec{L}_s$ curves, Phase I checks whether there is a point on $\uvec{L}_{o}$ that passes the desired local coordinates on the sphere $\bm{\Psi}_f$. If $\uvec{L}_o$ fails to reach $\bm{\Psi}_f$, $\zeta_q$ is re-tuned and kinematic model is resolved numerically. After succeeding Phase I, the curve is checked whether the final point of local coordinate $\bm{\Psi}_l$ is at desired values of $\bm{\Psi}_f$ as Fig. \ref{Fig:SPHERESECTION}. If the curve $\uvec{L}_o$ fails from $\bm{\Psi}_l=\bm{\Psi}_f$, the radius of desired virtual surface $R_q$ is increased by obtained error. Also, Phase II has a second extra step to reach the sphere final distance $\uvec{P}_l$ to the exact desired plane configuration $\uvec{P}_l = \uvec{P}_f$ by tuning $R_u$ and $\zeta_u$ parameters. In the final step, the final state as the desired spin angle $\psi_f$ is achieved. Note that this algorithm requires to numerically solve differential equations in $k$ iterations until full convergence of the chosen configuration.

\subsubsection{Phase I}
This phase solves the kinematic model (\ref{EQ:LatestStateEquation})-(\ref{Eq:AngleGoalthetaM}) first. Then, it finds the curve $\uvec{L}_o$ on the rotating sphere that passes $\bm{\Psi}_f$ while the sphere moves toward desired position $\uvec{P}_f$.

\begin{figure}[t!]
	\centering	
	\includegraphics[width=2.6 in,height=2.2 in]{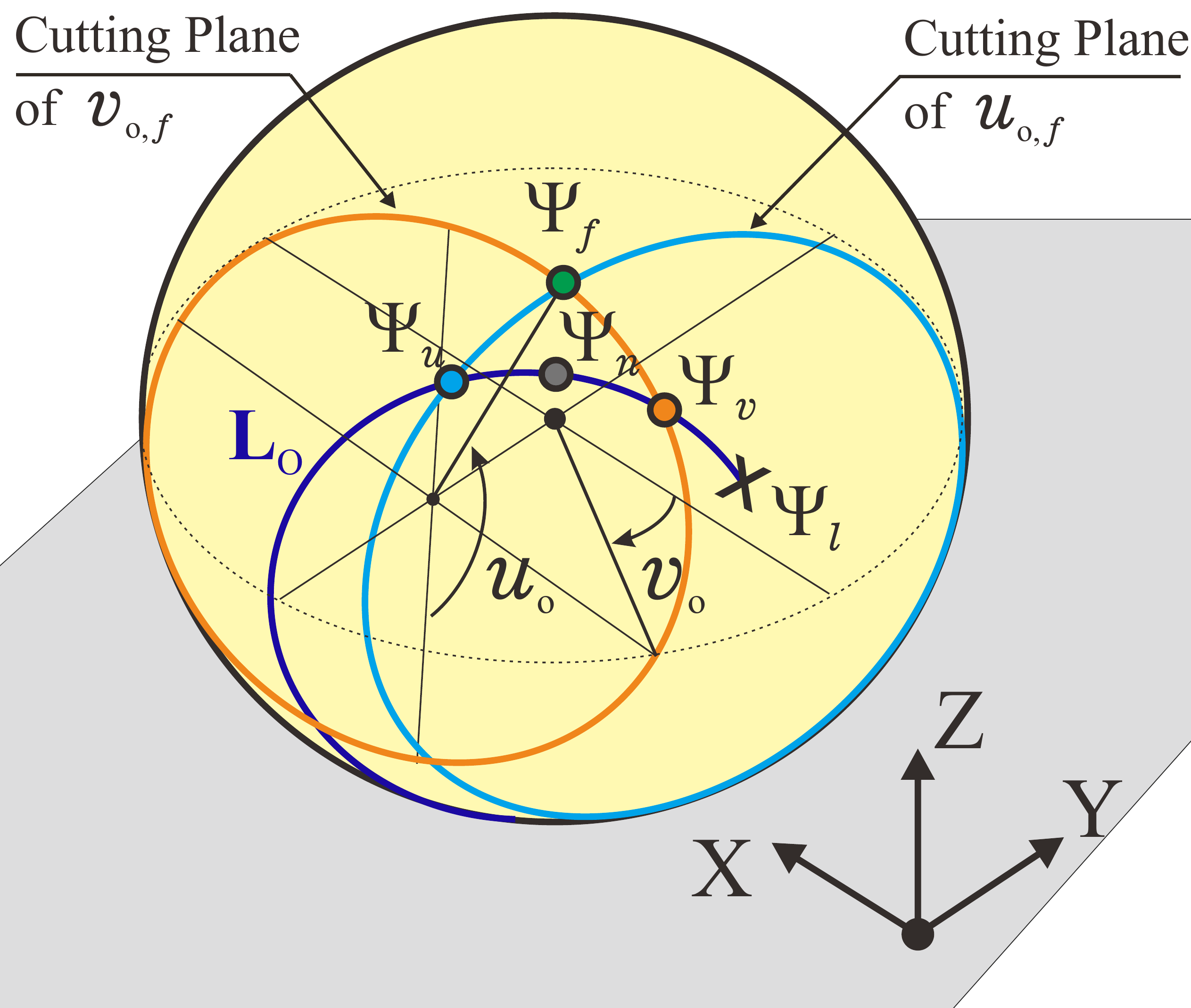}				
	\caption{Parametrization of the curve $\uvec{L}_{o}$ by cutting planes of $u_{o,f}$ and $v_{o,f}$.}\label{Fig:SPHERESECTION}
\end{figure}
Before defining the tuning functions, we have to develop a converging state for the spin angle. The spin motion can make tracking unstable when the sphere passes the same desired spin angle many times during its rotation. Thus, we develop an indirect convergence by using spinning angle deviation $\psi_q$ to converge the desired spin angle $\psi_f$. During calculations of minimum distance $d$ at (\ref{Eq:minimumditancerule}), we assumed that under-cap area $S_t$ of traversed distance $\uvec{L}_o$ is constant while $C_0 \rightarrow C_1$. Thus, the desired under-cap area is defined by $S_t=S'_c+R^2_{o}(\psi_f+\psi_u(k)-{\psi'})$ (only orange and blue cap areas excluding dashed-line part as Fig. \ref{Fig:SphereDistanceMotion2}), where $\psi_u(k)$ is the constant re-tuning spin angle for Phase III. Next, we extend the Gauss-Bonnet theorem between two points of $\{u_o(t),v_o(t),\psi(t)\}$ and $\{u_{o,f},v_{o,f},\psi_f\}$ in time $t$, as
\begin{equation}
\psi_q(t)=\frac{1}{R^2_o}\left[S_t-S_i(t)\right]= \frac{1}{R^2_o}\left[S_t- R^2_o\left(\psi(t)-\psi_0\right)\right],
\label{Eq:PureSpinAngleDEviation}
\end{equation}
where $S_i(t)$ and $\psi(t)$ are the changing cap area from the initial state till current time $t$ and the current spin angle.

As the last parameter in the kinematic model, the rolling rate (arc-length derivation relative to the time $\delta(t)$) is the parameter that is multiplied to all three inputs and drift term. This variable is related to the time ($ds/ dt$) and it varies the arc-length step of curve $\uvec{L}_{o}$ and $\uvec{L}_{s}$ in the given time $t$. Therefore, $\delta$ is defined to have a rest-to-rest motion by
\begin{equation}
\delta(t,u_o,u_s,v_s)=||\uvec{P}_f-\uvec{P}(t)||_2 \cdot \Big|   \frac{  v_{o,f} \cdot u'_o }{T} \Big|= \left[(u_{s,f}-u_s(t))^2+(v_{s,f}-v_s(t))^2\right]^{\frac{1}{2}} \cdot \Big| \frac{v_{o,f} \cdot u'_o }{T} \Big| ,
\label{Eq:DELTAEQuationtotal}
\end{equation}
where $T$ is the time scaling.

\alglanguage{pseudocode}
\begin{algorithm}[t!]
	\caption{Phase I}
	\begin{algorithmic}[1]
		\Procedure{ConfigSolve}{$\bm{\Psi}_f,\uvec{P}_f,\psi_f,\bm{\Psi}_0,\uvec{P}_0,\psi_0$}
		\While {$e_n \geq \epsilon_n$}
		\State Solve the kinematic model (\ref{EQ:LatestStateEquation}) numerically
		\State Collect $\bm{\Psi}_n$, $\bm{\Psi}_l$, $\bm{\Psi}_v$ and $\bm{\Psi}_u$
		\State Compute $e_n$ and  $e'_n$ by (\ref{EQ:ErrorNormalD})  \Comment{Error of nearest point to $\bm{\Psi}_f$}
		\If {$e_n(k)>\epsilon_n$}
		\State Apply the directional comparison by $\bm{\Psi}_n$, $\bm{\Psi}_v$ and $\bm{\Psi}_u$ \Comment{Details in \ref{RegionalComparison}}
		\State Calculate $\zeta_q(k)$ by (\ref{Eq:UpdateofZetaq})
		\EndIf
		\EndWhile
		\EndProcedure
	\end{algorithmic}
	\label{Algo:LoopComputationPhaseI}
\end{algorithm}
Algorithm~\ref{Algo:LoopComputationPhaseI} shows the computations of Phase I. By solving the Darboux-based kinematic model, the geometric parameters ([see Fig. \ref{Fig:SPHERESECTION}) are determined from obtained $\uvec{L}_{o}$ and $\uvec{L}_{s}$ trajectories. $\bm{\Psi}_l$ and $\uvec{P}_l$ are the final arrived configuration by the numerical solution of the found trajectories. Also, $\bm{\Psi}_n$, $\bm{\Psi}_u$ and $\bm{\Psi}_v$ are the nearest point on $\uvec{L}_{o}$ to $\bm{\Psi}_f$ and existing nearest points on $\uvec{L}_{o}$ that are obtained by cutting planes of $u_{o,f}$ and $v_{o,f}$ on the sphere. Note that all points are collected as the final existing values since the curve can passes the nearest point or cutting planes more than once.

In Algorithm~\ref{Algo:LoopComputationPhaseI}, $\epsilon_n$ is a small value for breaking while loop, when required accuracy is achieved for Phase I. Also, the error $e_n$ is the Euclidean distance of nearest $\bm\Psi_n$ and desired $\bm\Psi_f$ points in  $\mathbb{R}^3$
\begin{align}
\begin{split}
&e_n=||\bm{\Psi}_n-\bm{\Psi}_f||_3=R_{o}\big[(\sin v_{o,f} -\sin v_{o,n})^2+(\cos u_{o,f}\cos v_{o,f}-\cos u_{o,n} \cos v_{o,n})^2\\
&+(\sin u_{o,f} \cos v_{o,f}-\sin u_{o,n} \cos v_{o,n})^2\big]^{\frac{1}{2}}.
\end{split}
\label{EQ:ErrorNormalD}
\end{align}

Next, the absolute angle difference between $\bm{\Psi}_f$ and $\bm{\Psi}_n$ are utilized to update the $\zeta_q(k)$ angle in k-th iteration of Phase I as follows
\begin{equation}
\begin{split}
\zeta_q(k)=\zeta_q(k-1) \pm
\begin{cases}
& e'_{n} \cdot|Q^{zx}_f-Q^{zx}_n|,\;\;\;\;\;Q^{zx}_f=Q^{zx}_n\\
& e'_{n} \cdot|Q^{zy}_f-Q^{zy}_n|,\;\;\;\;\;Q^{zx}_f\neq Q^{zx}_n
\end{cases}
\end{split}
\label{Eq:UpdateofZetaq}
\end{equation}
where $e'_{n}=\min\{e_n\}$ is smallest error of $e_n$ till iteration $k$ and the angle differences with respect to $Z-X$ plane $Q^{zx}$ and $Z-Y$ plane $Q^{zy}$ for $\bm{\Psi}_f$/$\bm{\Psi}_n$ are obtained by Eq. (\ref{Eq:Coorinatequationsphereonplane}) as
\begin{align}
Q^{zx}=\left| \tan^{-1}\left(\frac{\sin u_o \cos v_o}{\cos u_o \cos v_o}\right)\right|=|u_o|, Q^{zy}=\left|\tan^{-1}\left(\frac{\sin u_o \cos v_o}{\sin v_o}\right)\right|.
\label{Eq:TheQCOFormulas}
\end{align}
The sign of each update in (\ref{Eq:UpdateofZetaq}) is chosen from the directional updates in \ref{RegionalComparison}.  The goal of the directional update is to always move the curve $\uvec{L}_o$ towards $\bm{\Psi}_{f}$ on the spherical surface $U_C$.
\subsubsection{Phase II and III}

\alglanguage{pseudocode}
\begin{algorithm}[t!]
	\caption{Complete Computation}
	\begin{algorithmic}[1]
		\Procedure{ConfigCom}{$\bm{\Psi}_f,\uvec{P}_f,\psi_f,\bm{\Psi}_0,\uvec{P}_0,\psi_0$}
		\While {$e_s \geq \epsilon_s $} \Comment{Phase III }
		\While {$ \left(e_n \geq \epsilon_n\right) \;\& \;\left(e_r \geq \epsilon_r\right) \;\&\; \uvec{P}_l \neq \uvec{P}_f\;$} \Comment{Phase II}
		\While {$e_r \geq \epsilon_r$} \Comment{Phase II}
		\State C{\footnotesize ONFIG}S{\footnotesize OLVE}(.) \Comment{Phase I}
		\State Compute $e_r$, $e'_r$, $n_s$ and $d_s$
		\If {$\bm{\Psi}_l \neq \bm{\Psi}_n$} \Comment{Updates $R_q$}
		\State $R_q(k)\leftarrow+R_q(k-1)$ according to (\ref{Eq:Rupdateofphase2})
		\Else
		\State $R_q(k)\leftarrow-R_q(k-1)$ by $n_s\leq 1$ case according to (\ref{Eq:Rupdateofphase2})
		\EndIf
		\EndWhile
		\If {$||\uvec{P}_f-\uvec{P}_l||_2>\epsilon_p$} \Comment{Phase II}
		\State Update $R_u$ and $\zeta_u$
		\EndIf
		\EndWhile
		\If {$\psi_l\neq\psi_f$} \Comment{Phase III}
		\State Update $\psi_u$
		\EndIf
		\State Reset $e_n$, $e'_n$, $e_r$ and $e'_r$
		\EndWhile\\
		\Return $\uvec{L}_o$ and $\uvec{L}_s$
		\EndProcedure
	\end{algorithmic}
	\label{Algo:Loop2Computation}
\end{algorithm}
After finding the suitable curve $\uvec{L}_o$ from Phase I that passes $\bm{\Psi}_f$, Phase II tunes variable for converging $\uvec{L}_o$ and $\uvec{L}_s$ final arrived points ($\bm{\Psi}_l$ and $\uvec{P}_l$) to the desired configurations on $\bm{\Psi}_f$ and $\uvec{P}_f$. Algorithm \ref{Algo:Loop2Computation} depicts the overall computations, including all the phases using the flowchart of Fig. \ref{Fig:BLOCKDIAGRAMFeedForward}-a.

In Phase II, the error $e_r$ of final arrival point $\bm{\Psi}_l$ with respect to desire values of $\bm{\Psi}_f$ is first determined by
\begin{align}
\begin{split}
&e_r=||\bm{\Psi}_l-\bm{\Psi}_f||_3=R_{o}\big[(\sin v_{o,f} -\sin v_{o,l})^2+(\cos u_{o,f}\cos v_{o,f}-\cos u_{o,l}\cos v_{o,l})^2\\
&+(\sin u_{o,f} \cos v_{o,f}-\sin u_{o,l}\cos v_{o,l})^2\big]^{\frac{1}{2}}.
\end{split}
\label{Eq:ErrorERAlgo}
\end{align}
Next, the update of $R_q$ is to enlarge the loops of $\uvec{L}_o$ for shorting the distance between nearest $\bm\Psi_n$ and final arrived $\bm{\Psi}_l$ points; hence, the algorithm computes the $R_q$ update for $k$-th iteration by using the error $e_r$ as
\begin{equation}
\begin{split}
R_q(k)= R_q(k-1)+ R_o \cdot
\begin{cases}
&  d_s \cdot e'_r,  \;\;\;\;\;\;\;\;\;\;\;\;\;\;\;\;\;\;\;\; n_s \leq 1 \\
&  n_s \cdot d_s \cdot e'_r,  \;\;\;\;\;\; \;1< n_s \leq 2 \\
& \left( d_s \cdot e'_r \right) / n_s,   \;\;\;\;\;\;\;\;\;\;\; n_s > 2
\end{cases}
\end{split}
\label{Eq:Rupdateofphase2}
\end{equation}
where $e'_r=\min\{e_r\}$, $d_s$ and $n_s$ are the smallest error of $e_r$ till iteration $k$,  distance ratio of $\uvec{L}_s$ for the loop on the curve $\uvec{L}_o$ and the numbers of created full loops by spinning $\psi$, respectively. The distance ratio $d_s $ and the number of created loops $n_s \in \mathbb{R}$ are calculated as follows
\begin{equation}
d_s=\frac{||\uvec{P}_n-\uvec{P}_f||_2}{||\uvec{P}_0-\uvec{P}_f||_2}=\left[\frac{(u_{s,n}-u_{s,f})^2+(v_{s,n}-v_{s,f})^2}{(u_{s,0}-u_{s,f})^2+(v_{s,0}-v_{s,f})^2}\right]^{\frac{1}{2}}, n_s=|\psi_l/\psi_n|.
\end{equation}
Note that $n_s$ is computed with the assumption that $\psi$ value can be greater than $2\pi$ which every $2\pi $ orientation represents a loop on spin angle \cite{tapp2016differential}.

As the final step of Phase II, because our arc-length-based inputs create curve $\uvec{L}_o$ with multiple loops (specially for $d \ll ||\uvec{P}_f-\uvec{P}_0||_2$), we use $R_u$ and $\zeta_u$ variable for shifting $\uvec{L}_o$ away from $\bm\Psi_0$ on the surface of the sphere $U_C$ (can be looked as decreasing the size of the enlarged $R_q$) with same expressed flexible rope model. This computation step with updating $\zeta_u$ and $R_u$ tries to reach $\bm\Psi_l \rightarrow \uvec{P}_f$ as previous loops try to keep $\bm\Psi_l=\bm\Psi_f$ condition true,
\begin{align}
\begin{split}
&R_u(k)=R_u(k-1)-R_q(k) \cdot \frac{||\uvec{P}_l-\uvec{P}_f||_2}{||\uvec{P}_l-\uvec{P}_0||_2},\\
&\zeta_u(k)=\zeta_u(k-1)+\begin{cases}
&-\tan^{-1}\left(\frac{R_u(k)\tan \zeta_q (k-1) }{R_o}\right), \;\;\;\;\;\textnormal{ for } v_{o,f} \geq 0\\
&\tan^{-1}\left(\frac{R_u(k)\tan \zeta_q (k-1) }{R_o} \right), \;\;\;\; \;\;\;\;\textnormal{ for } v_{o,f}<0
\end{cases}
\end{split}
\label{Eq:TuningPlanePhaseII}
\end{align}
Final phase is to update the tuning variable of the spinning angle $\psi_u$ in Eq. (\ref{Eq:PureSpinAngleDEviation}) with
\begin{equation}
\psi_u(k)=\psi_u(k-1)-e'_s\sign{\left[\psi_f-\psi_l\right]},
\end{equation}
where $e'_s=\min\{\psi_f-\psi_l\}$ is the smallest error till iteration $k$. Also, $\epsilon_s$, $\epsilon_r$ and $\epsilon_p$ are the accuracies of each designated step in Algorithm \ref{Algo:Loop2Computation}.

\section{Global Convergence of The Planning Algorithm}
To prove the convergence of proposed iterative Algorithm \ref{Algo:Loop2Computation}, one can utilize Zangwill's convergence theorem  \cite{zangwill1969nonlinear}.
\begin{thm}
	Let the iterative algorithm $\uvec{A}: \; \mathbb{R}^5 \rightarrow \mathbb{R}^3$, showed in the flowchart at Fig. \ref{Fig:BLOCKDIAGRAMFeedForward}, be on $\uvec{X}=\uvec{X}_I \circ \uvec{X}_{II} \circ \uvec{X}_{III}$ compact set where $\uvec{X}_I$, $\uvec{X}_{II}$ and $\uvec{X}_{III}$ are sets of phase I, II and phase III computation steps, respectively. Given $\uvec{x}_0 \in \uvec{X}$, the created sequence $\{\uvec{x}_k\}^{\infty}_{k=1}$ satisfies
	$\uvec{x}_{k+1} \in \uvec{A}(\uvec{x}_k)$. Then, $\uvec{A}$ is globally convergent to a solution $\bm{\Gamma} \subset \uvec{X}$ with following conditions
	\begin{align*}
	\begin{split}
	& \bullet \exists\textit{ a descent function $\uvec{z}$ for }\bm\Gamma\textit{ and }\uvec{A}, \\
	&\bullet \textit{The squence}\; \{\uvec{x}_k\}^{\infty}_{k=0}\subset S\;\textnormal{for}\; S \subset \uvec{X} \textit{ is a compact set,}  \\
	&\bullet \textit{The mapping } \uvec{A}\textit{ is closed at all points of } \uvec{X}/\bm\Gamma.
	\end{split}
	\end{align*}	
	\label{TheoremGlobalConvergence}
\end{thm}

Regarding the first condition in Theorem \ref{TheoremGlobalConvergence}, the designed line search algorithm $\uvec{A}$ have descent function $\uvec{z}=\{R_q, \zeta_q,R_u, \zeta_u,\psi_u\}$ which roughly can be presented as
\begin{equation}
\uvec{z}(x)=\uvec{x}_k+\uvec{h}_k\uvec{d}_k
\label{Eq:desentfunction}
\end{equation}
where $\uvec{d}_k$ and $\uvec{h}_k=\textnormal{diag}\{h(1),h(2),h(3),h(4),h(5)\}$  are the direction of iteration and step size of iterations. Note that $\zeta_u$ is updated in the same way with others, only because $R_u$ changes the curve $\uvec{L}_o$, the new angular location shift happens by $\zeta_u$. Next, the given descent function $\uvec{z}(x)$ has following properties:
\begin{align*}
\begin{cases}
&\textnormal{If } \uvec{x}\notin\bm\Gamma\textnormal{ and } \uvec{y}\in \uvec{X},\;\uvec{z}(y)<\uvec{z}(x), \\
&\textnormal{If } \uvec{x}\in\bm\Gamma\textnormal{ and } \uvec{y}\in \uvec{X},\;\uvec{z}(y)\leq\uvec{z}(x),
\end{cases}
\end{align*}	
hence, the first condition of Theorem \ref{TheoremGlobalConvergence} is satisfied. The algorithm $\uvec{A}$ is on the compact manifold of the sphere $U_C$ and the plane $U_S$ so $\uvec{h}_k$ norms as the error e.g., $e_r$, $\mathbb{R}^n \rightarrow \mathbb{R}$ become a compact set as well. This can be clearly extended for the second condition about all produced sequences $\{\uvec{x}_k\}^{\infty}_{k=0}$ on $\uvec{X}$. Now, we provide a proposition to satisfy final condition as

\begin{prop}
	Let $f$ be a real continues function on $\uvec{X}$. Then, Algorithm $\uvec{A}$ with solution set $\bm\Gamma$ of
	\begin{equation*}
	\bm\Gamma(\uvec{x},\uvec{d})=\{\uvec{y}\in \mathbb{R}^n\; | \;\uvec{y}=\uvec{x}+\uvec{h} \uvec{d},\; \uvec{h}>0, \; \textnormal{and } f(\uvec{y})= \min{\;f(\uvec{x}+\uvec{h} \uvec{d})}     \}
	\end{equation*}
	is closed at any point $(\uvec{x},\uvec{d})$ at which always $\uvec{d} \neq 0$.
\end{prop}
\ref{ProofofClosedSet} presents the proof of the given proposition. By satisfying all the three conditions in Theorem \ref{TheoremGlobalConvergence}, the proposed line search algorithm is always convergent to a solution for the desired configuration.
\begin{figure}[t!]
	\centering	
	\includegraphics[width=5 in, height= 2.2 in]{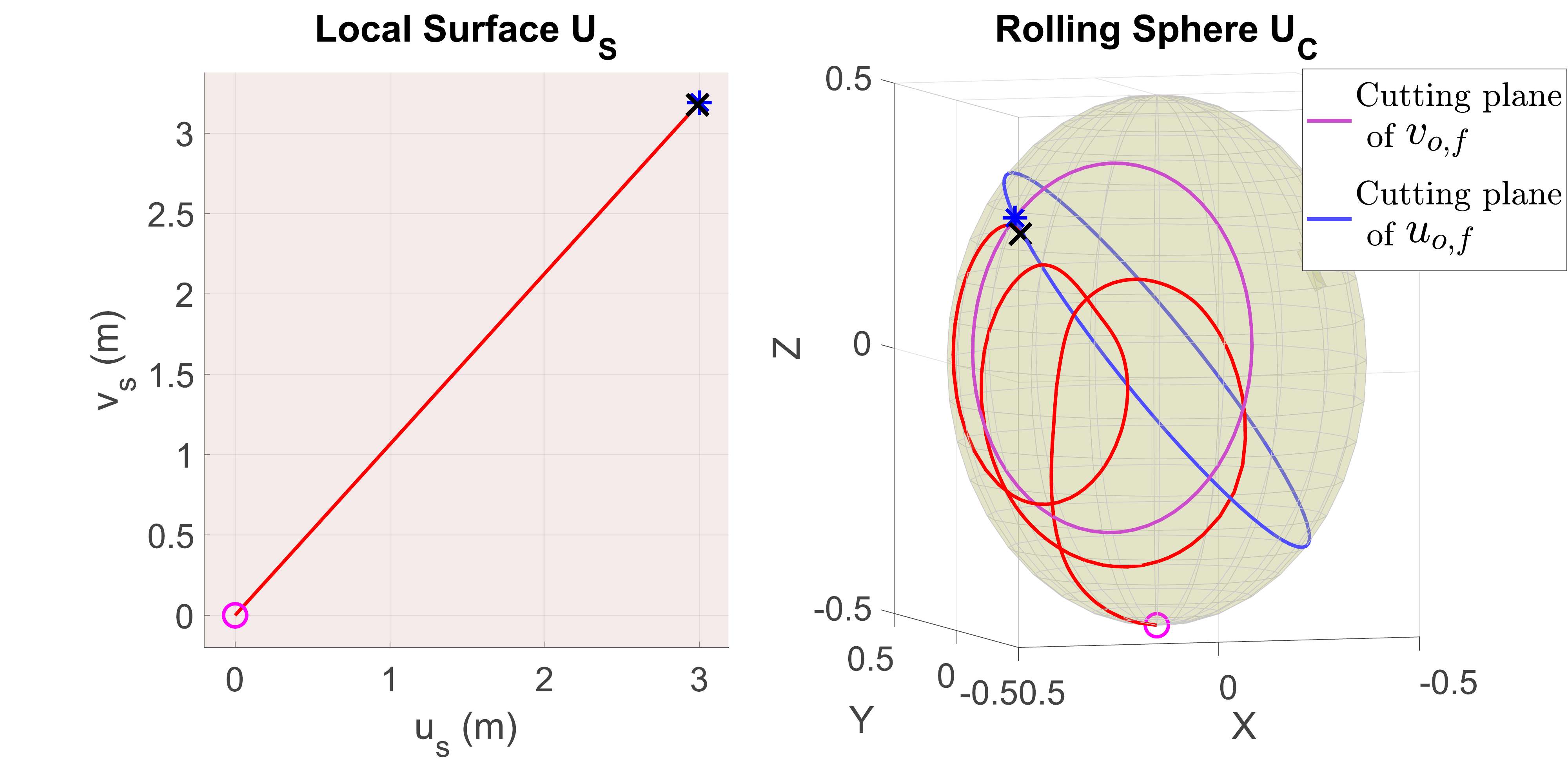}
	\includegraphics[width=3.4 in, height= 3.1 in]{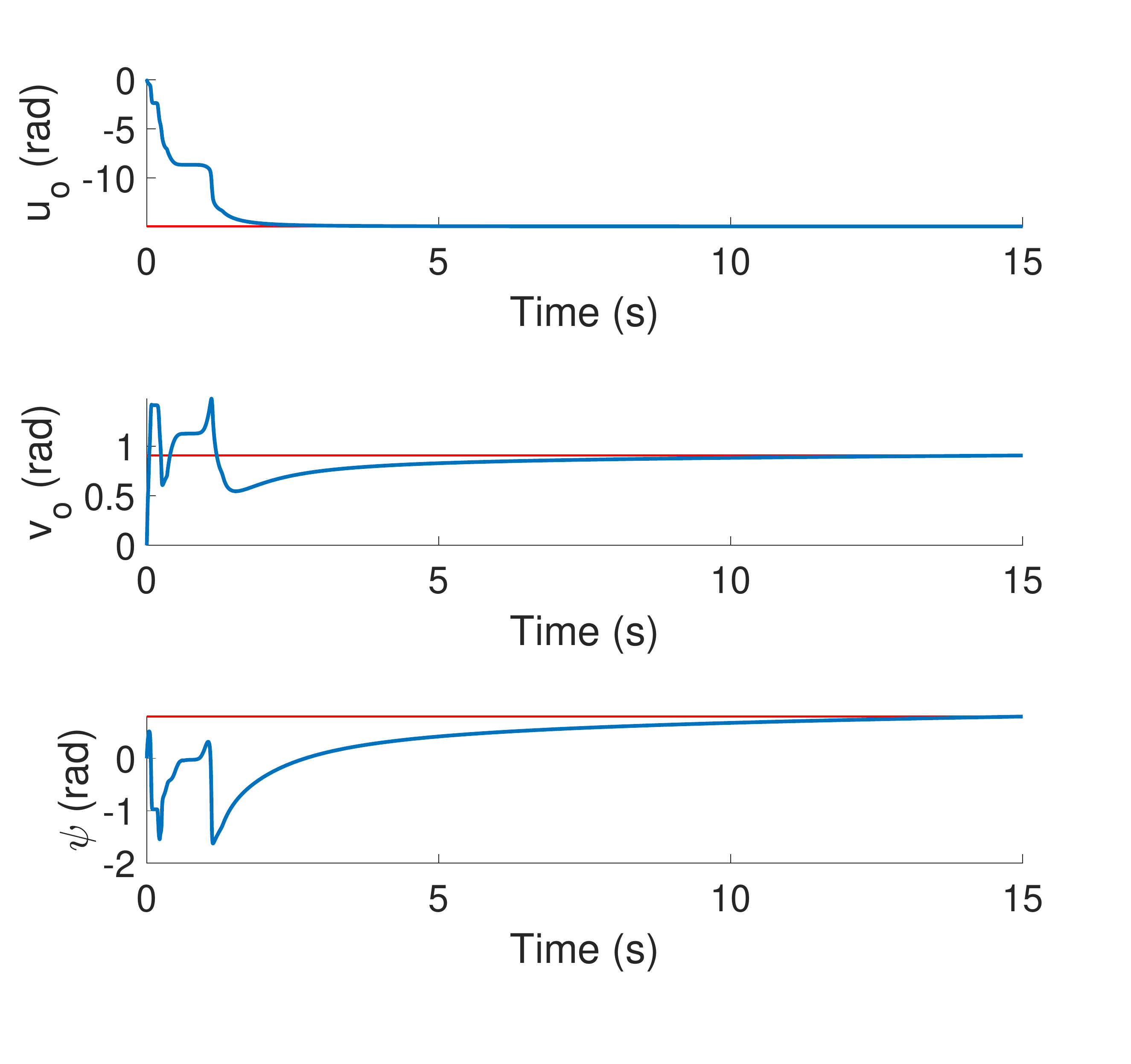}		
	\caption{Simulation results for a case study with the final configuration $\{u_{s,f},v_{s,f},u_{o,f},v_{o,f},\psi_f\}=\{3,3.2,-\frac{\pi}{2}-0.8,0.8,0.8\}$.}\label{Fig:Case1A1}
\end{figure}

\section{Results and Discussion}
The operation and performance of the proposed planning approach are tested under simulations. First, we check how phases of the algorithm work to achieve a successful convergence. Next, we analyze the simulation results for different desired spin angles while the rest of the configuration is the same.

\begin{figure}[t!]
	\centering	
	1. \includegraphics[width=4.05 in]{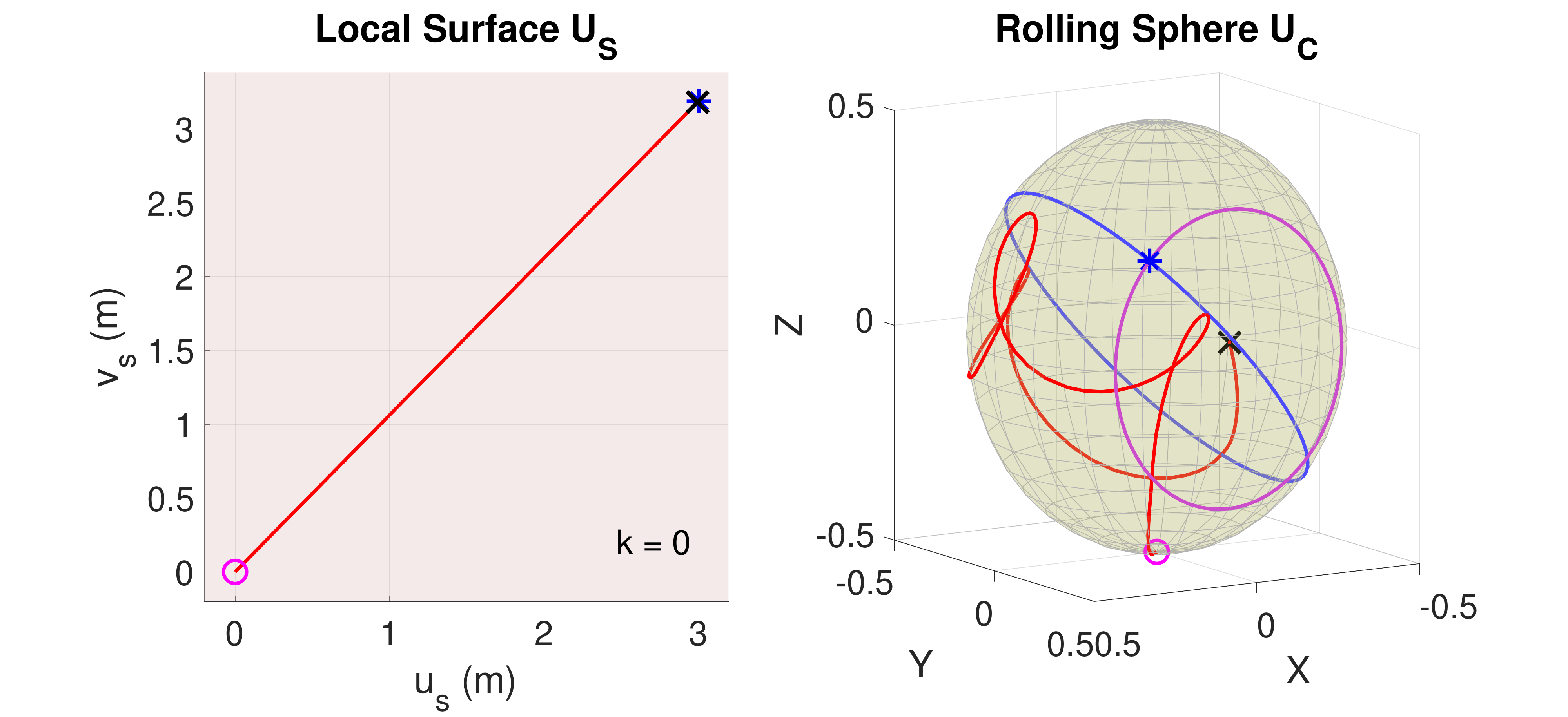}		\\	
	2. \includegraphics[width=4.05 in]{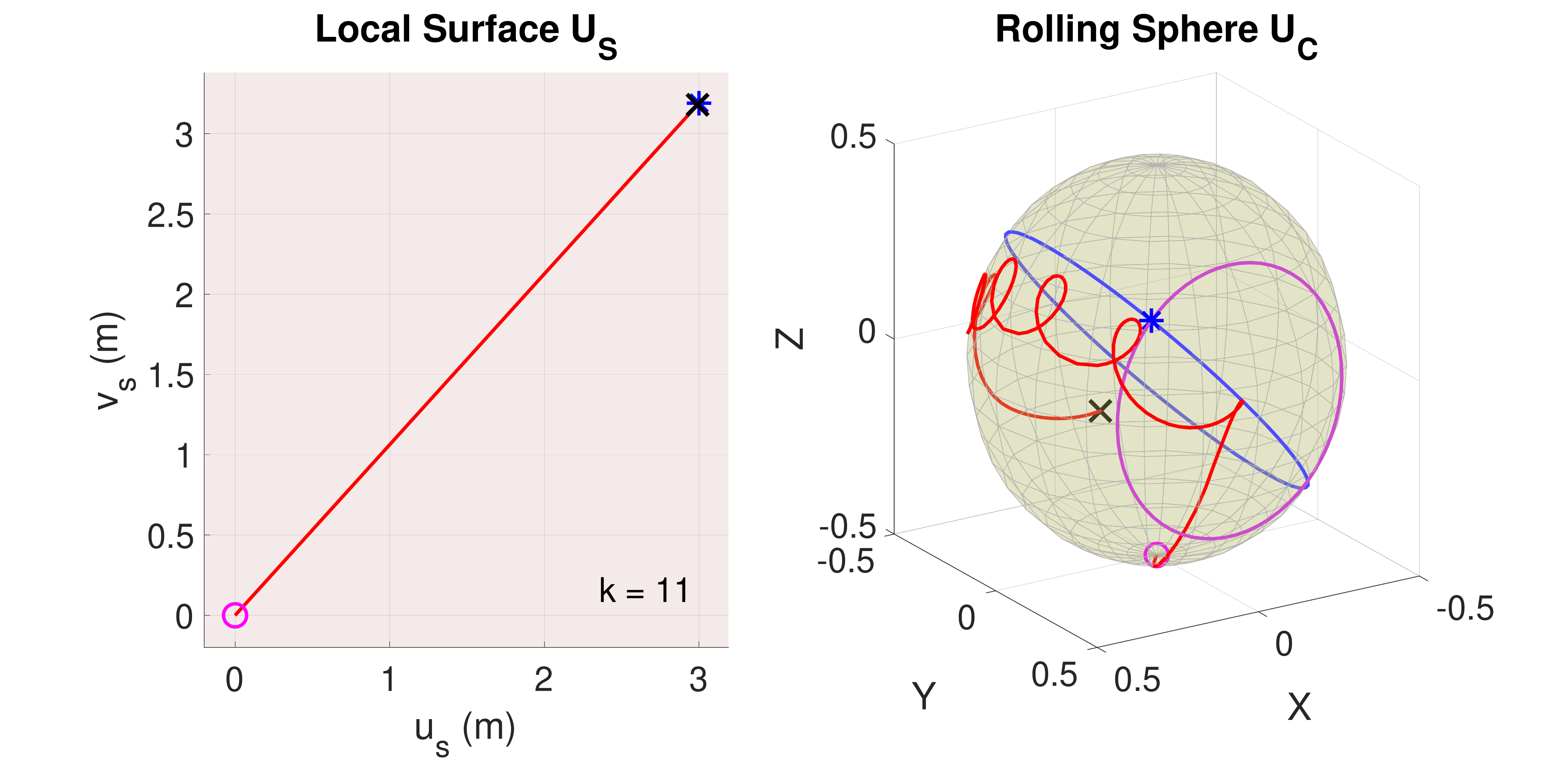}\\
	3. \includegraphics[width=4.05 in]{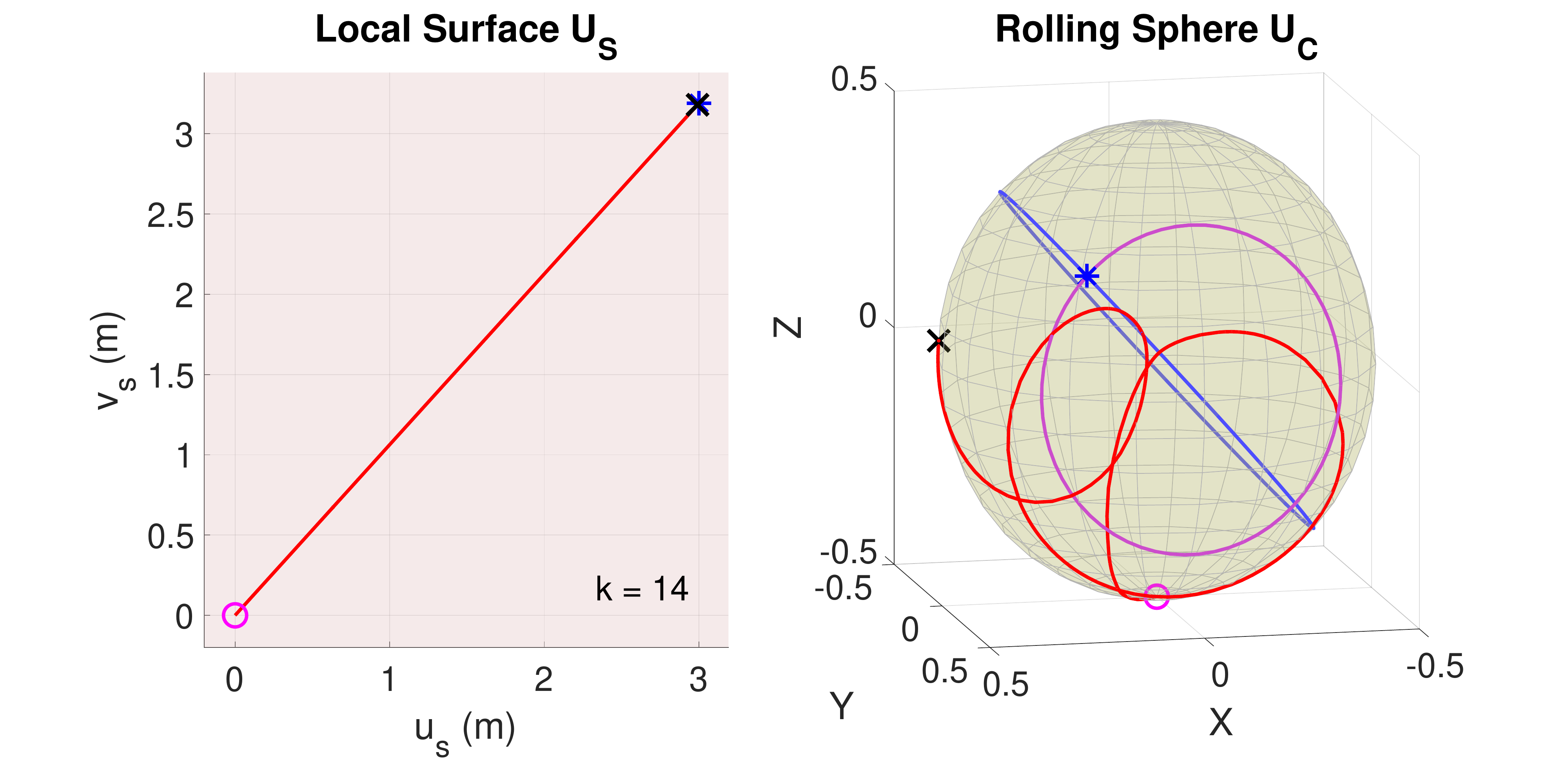}	\\
	4. \includegraphics[width=4.05 in]{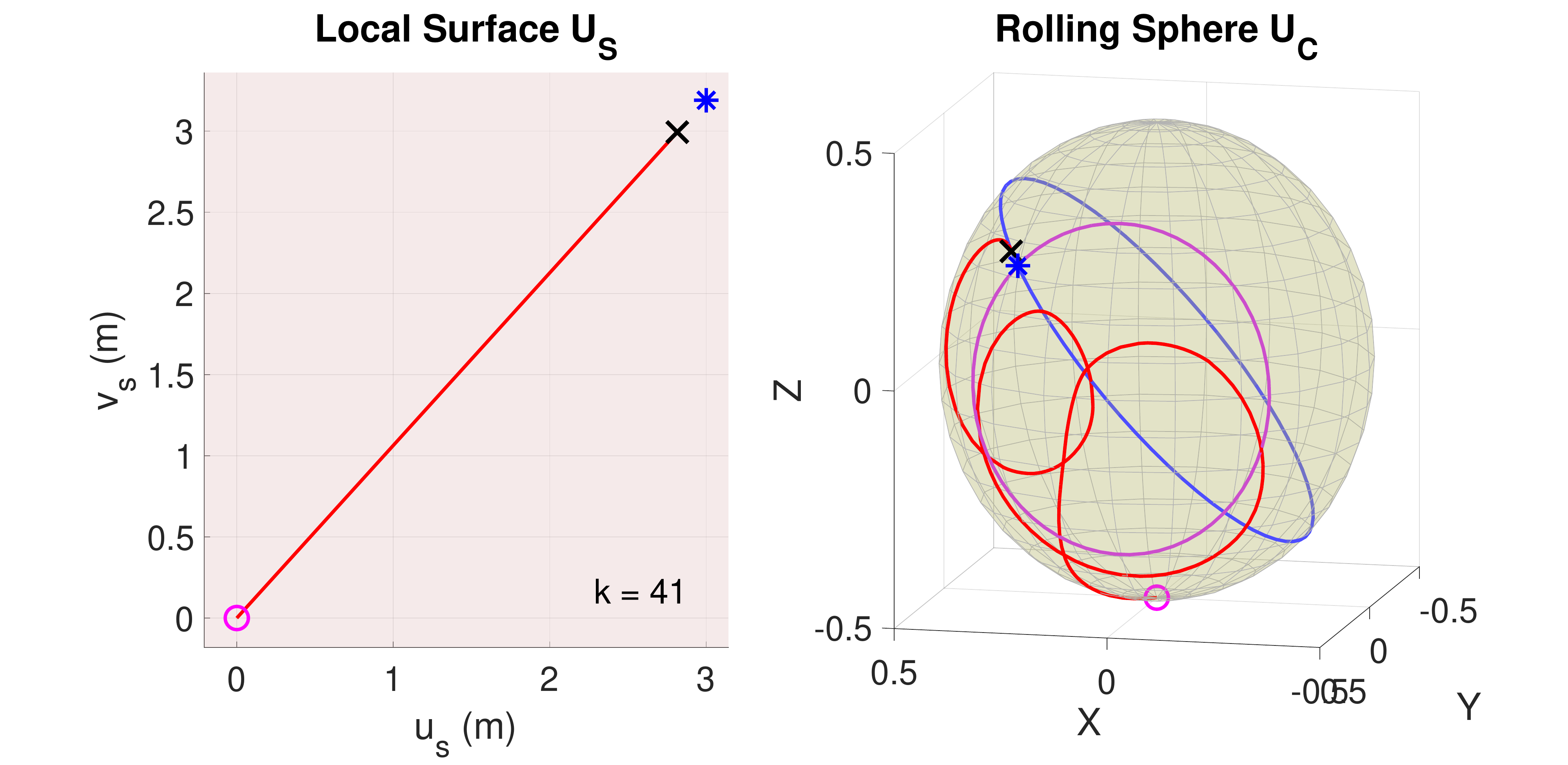}		
	\caption{Collected iterations of the given final configuration: 1. The first solution of system ($k=0$), 2. The first successful results of Phase I ($k=11$), 3. Converging Phase II ($k=14$), 4. The first results before applying Phase III ($k=41$). }\label{Fig:fullview}
\end{figure}

In the simulations, the kinematic model (\ref{EQ:LatestStateEquation})-(\ref{Eq:AngleGoalthetaM}) is solved together with (\ref{EQ:VirtualSurfaceGeodesicinputmain}) and (\ref{Eq:DELTAEQuationtotal}) by using Matlab's ODE45 routine.
We set the initial and final configurations as $\{u_{s,0}$,$v_{s,0}$,$u_{o,0}$,$v_{o,0}$,$\psi_0\}=\{0$,$0$,$0$,$0$,$0\}$ and $\{u_{s,f}$, $v_{s,f}$, $u_{o,f}$, $v_{o,f}$, $\psi_f$$\}=\{3$, $3.2$, $-\frac{\pi}{2}-0.8$, $0.8$ ,$0.8\}$. Note that our plane's final configuration satisfies distance constraint (\ref{Eq:minimumditancerule}) where $d$ for our case is $2.15\leq||\uvec{P}_{f}-\uvec{P}_{0}||_2$, and to see curve with multiple loops, we have $||\uvec{P}_{f}-\uvec{P}_{0}||_2= 4.38 $ m. For this case, the simulation time is set to $t_f=15$ s and time constant $T$ in (\ref{Eq:DELTAEQuationtotal}) is 1.
The accuracy of phases in Algorithm \ref{Algo:Loop2Computation} are $ \epsilon_n= \epsilon_r=0.07$, $ \epsilon_p=0.12$ and $ \epsilon_s=0.05$. As expressed in the controllability analysis in Ref. \cite{Tafrishi2021DarKiF}, there is a uncontrollable point at $\pi/4$ on the plane. For this case, we chose it away from point with $G_f= 0.754 $ rad to see our approach abilities. The initial value of $R_q$ is 0.005 to prevent any computation singularities. Also, we set the sphere radius and maximum divider as $R_o=0.5$ m and $\mu_r=4$.
\begin{figure}
	\centering			
	\includegraphics[width=2.9 in]{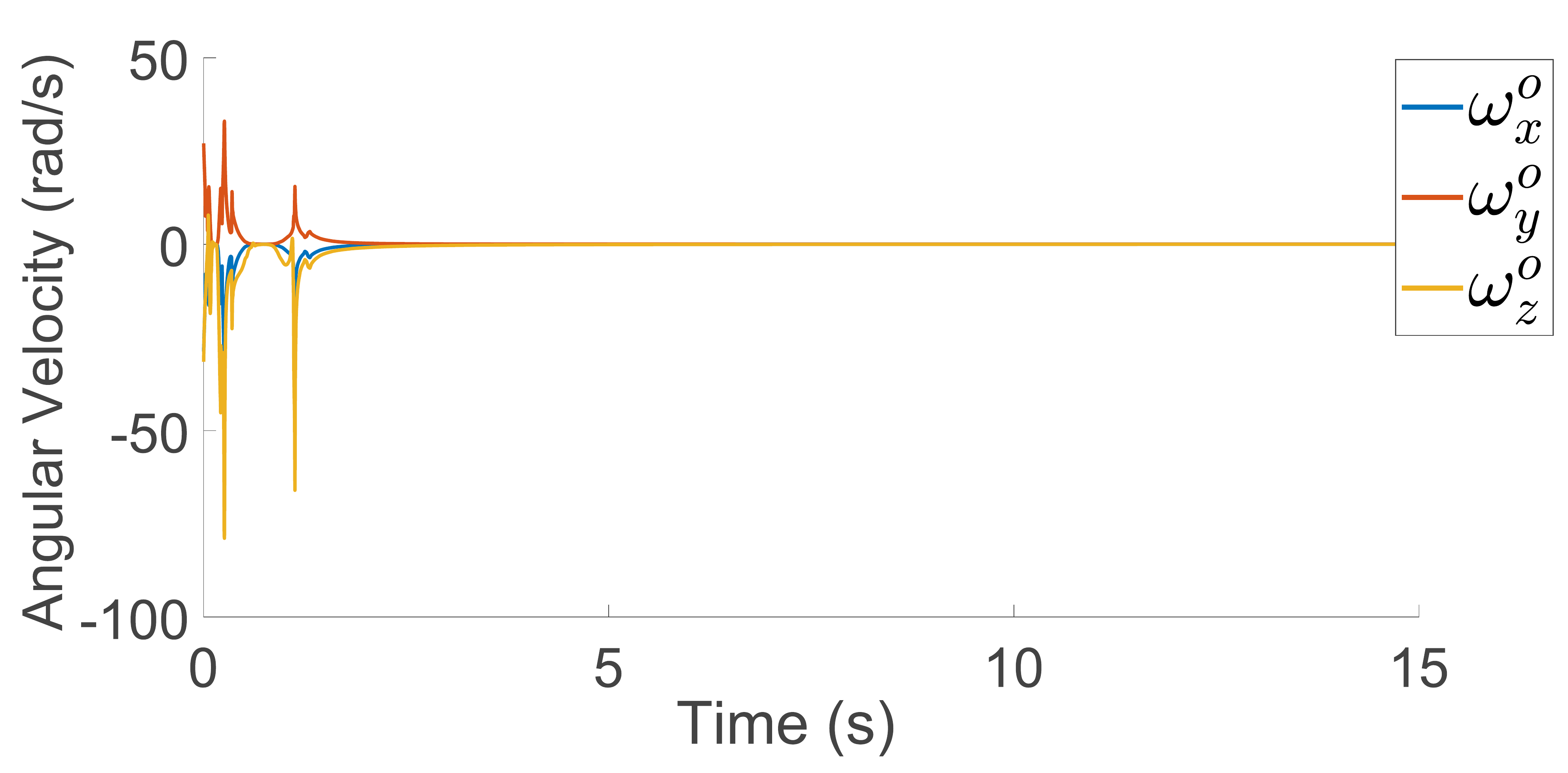}	
	\caption{ Angular velocities of the sphere with the designed arc-length-based based controller.} \label{Fig:Angular_Result1}
\end{figure}

After running the simulation, the final results are shown as Fig. \ref{Fig:Case1A1}.
The video of the simulation results for this section is available on Youtube \cite{amirhorde_2021}.
The calculation steps $k$ for this case is achieved with 48 iterations. The computation process takes about 7.2 s. By the achieved successful final iteration, we can interpret that while the sphere is rolling along the given desired final position $\uvec{P}_f$, it spins with smooth trajectories toward its final configuration for $\{\bm{\Psi}_f,\psi_f\}$.
\begin{figure}
	\centering		
	a) \includegraphics[width=2.7 in, height = 1.5 in]{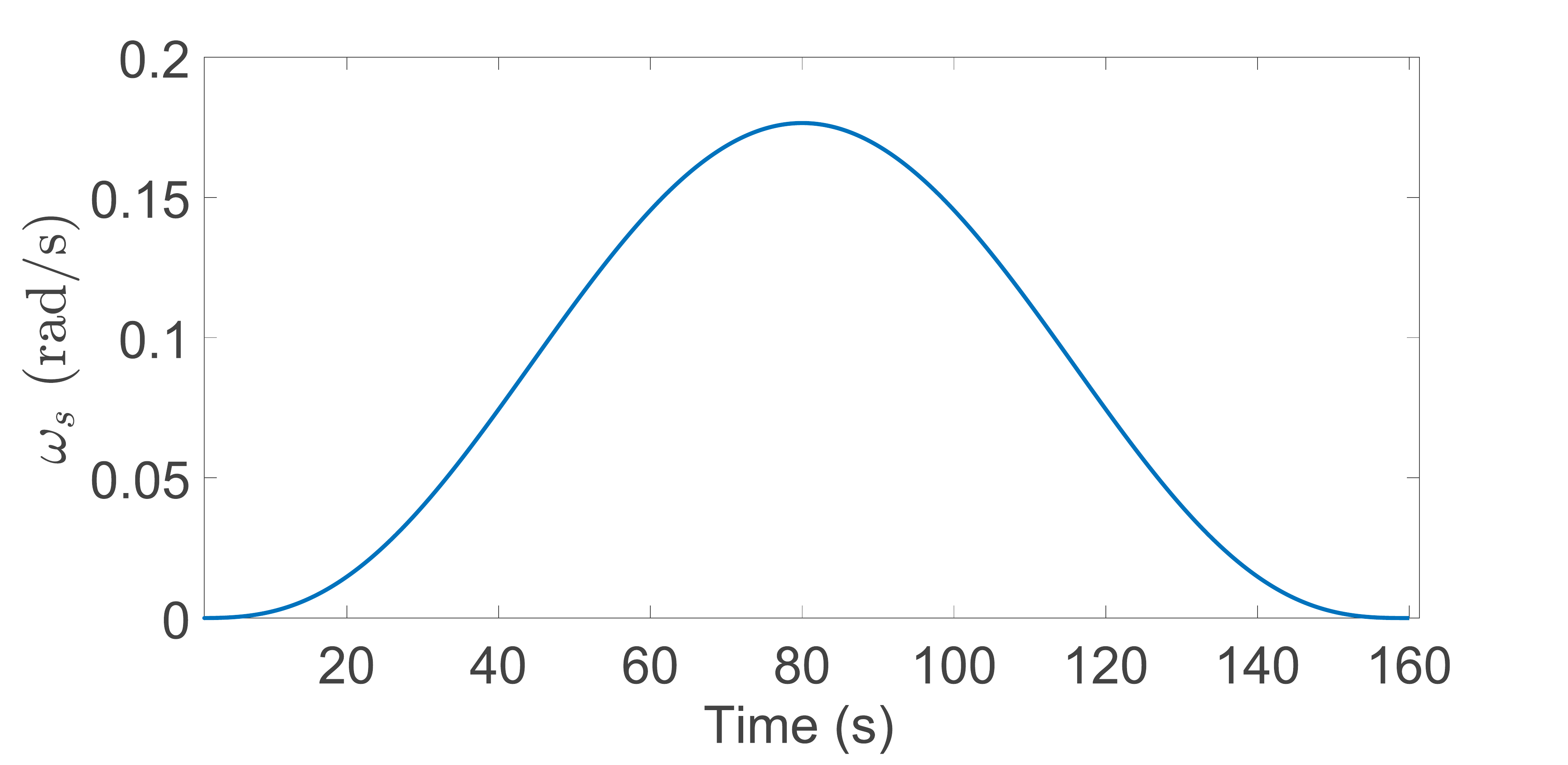} 		
	b) \includegraphics[width=3.3 in, height = 1.6 in]{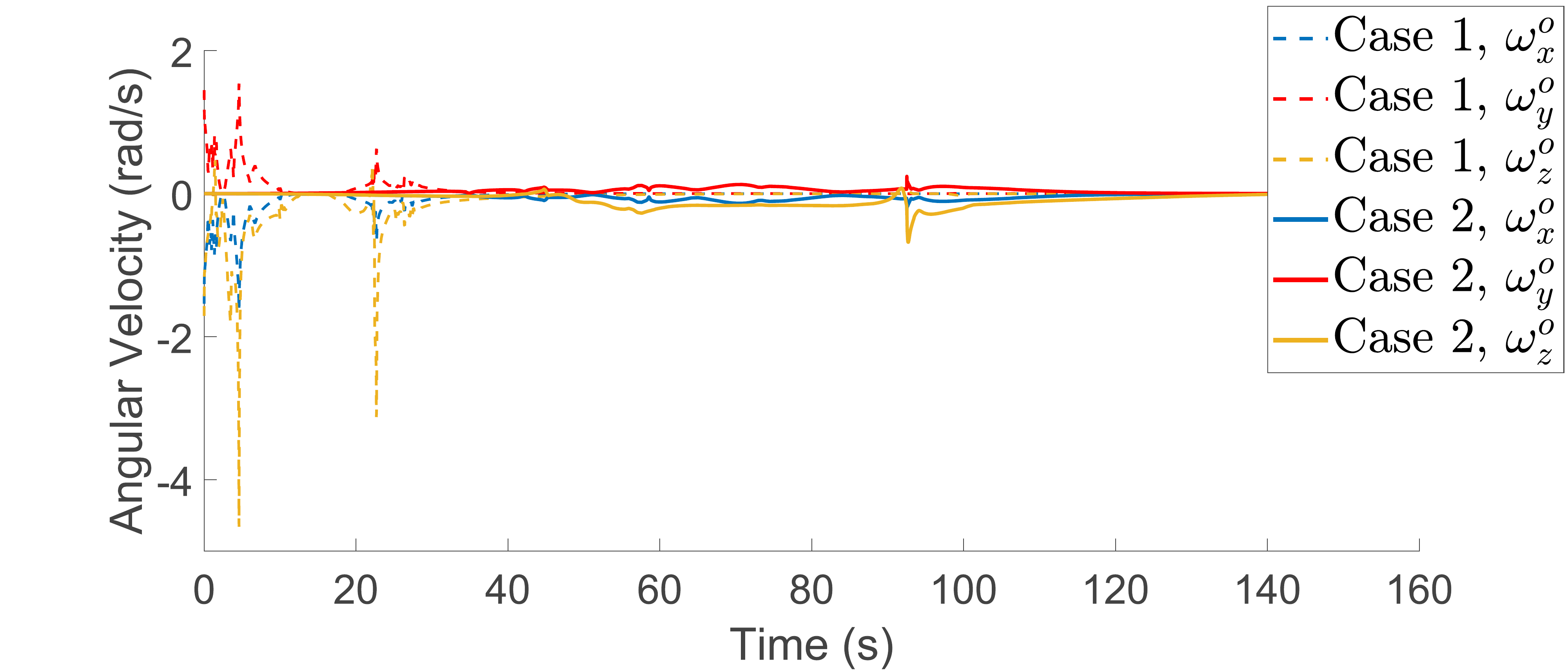}	
	\caption{ a) The designed function $\omega_{s}$ in time, b) Smoothed angular velocities of the sphere in two different cases of time scaling $T$.} \label{Fig:Main_Input_OMEGA1}\label{Fig:Beta}
\end{figure}

To understand better how the algorithm works by utilizing the derived virtual surface, we plot four collected iteration steps [see Fig. \ref{Fig:fullview}]. From the initial iteration ($k=0$), it is clear that $\uvec{L}_o$ is created by the designed virtual surface toward the desired states $\bm{\Psi}_f$ but because $\uvec{P}_f$ is far, $\uvec{L}_o$ curve moves along the $u_{o,f}$ cutting plane. A rope on a sphere explains the way that the achieved curve $\uvec{L}_o$ on $U_C$ can be imagined. So, the algorithm tunes the virtual surface and this bends the obtained curve $\uvec{L}_o$ in loops, like a rope, toward the desired $\bm\Psi_f$. Here, Phase I rises the $\uvec{L}_o$ to move $\bm\Psi_n$ near to $\bm\Psi_f$ as the result of succeeded iteration $k=11$. To move $\bm\Psi_l$ on $\bm\Psi_f$, Phase II enlarges the radius of the rope-like curve $\uvec{L}_o$, as the example iteration $k=14$ shows. The iteration $k=41$ in Fig.  \ref{Fig:fullview} satisfies $\bm\Psi_l=\bm\Psi_f$ condition while the next step is tuning the plain configuration to achieve $\uvec{P}_l=\uvec{P}_f$ using Eq. (\ref{Eq:TuningPlanePhaseII}). Finally, the Phase III condition is satisfied for the desired $\psi_f$ value shown in Fig. \ref{Fig:Case1A1}.
\begin{figure}[t!]
	\centering			
	\includegraphics[width=3.2 in, height= 2.8 in]{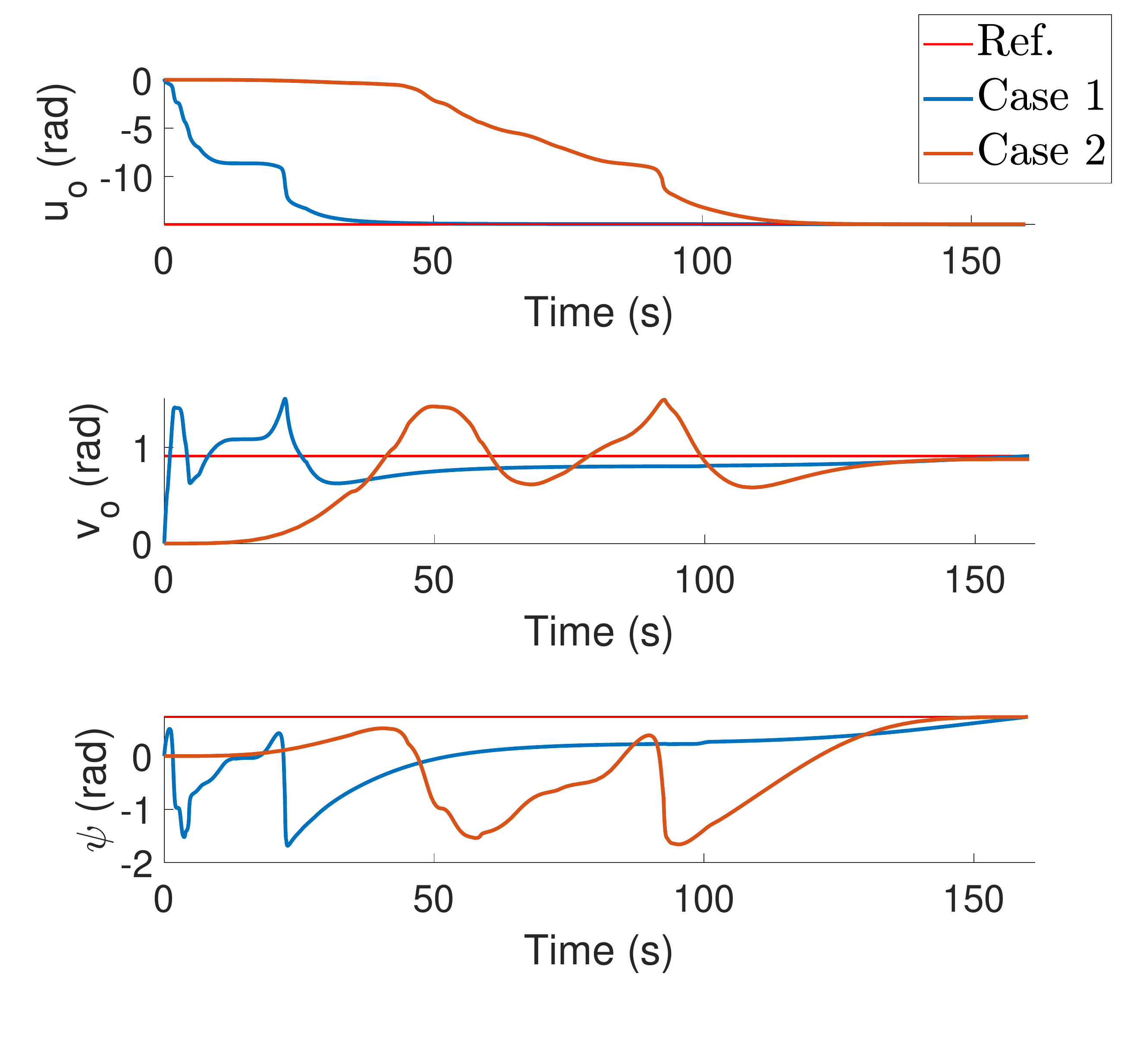}	
	\caption{ Comparing angular convergence of $\bm{\Psi}_f$ at two cases. }\label{Fig:Angular_Vel_com_pos1}
\end{figure}
\begin{figure}[t!]
	\centering	
	\includegraphics[width=5 in, height= 2.6 in]{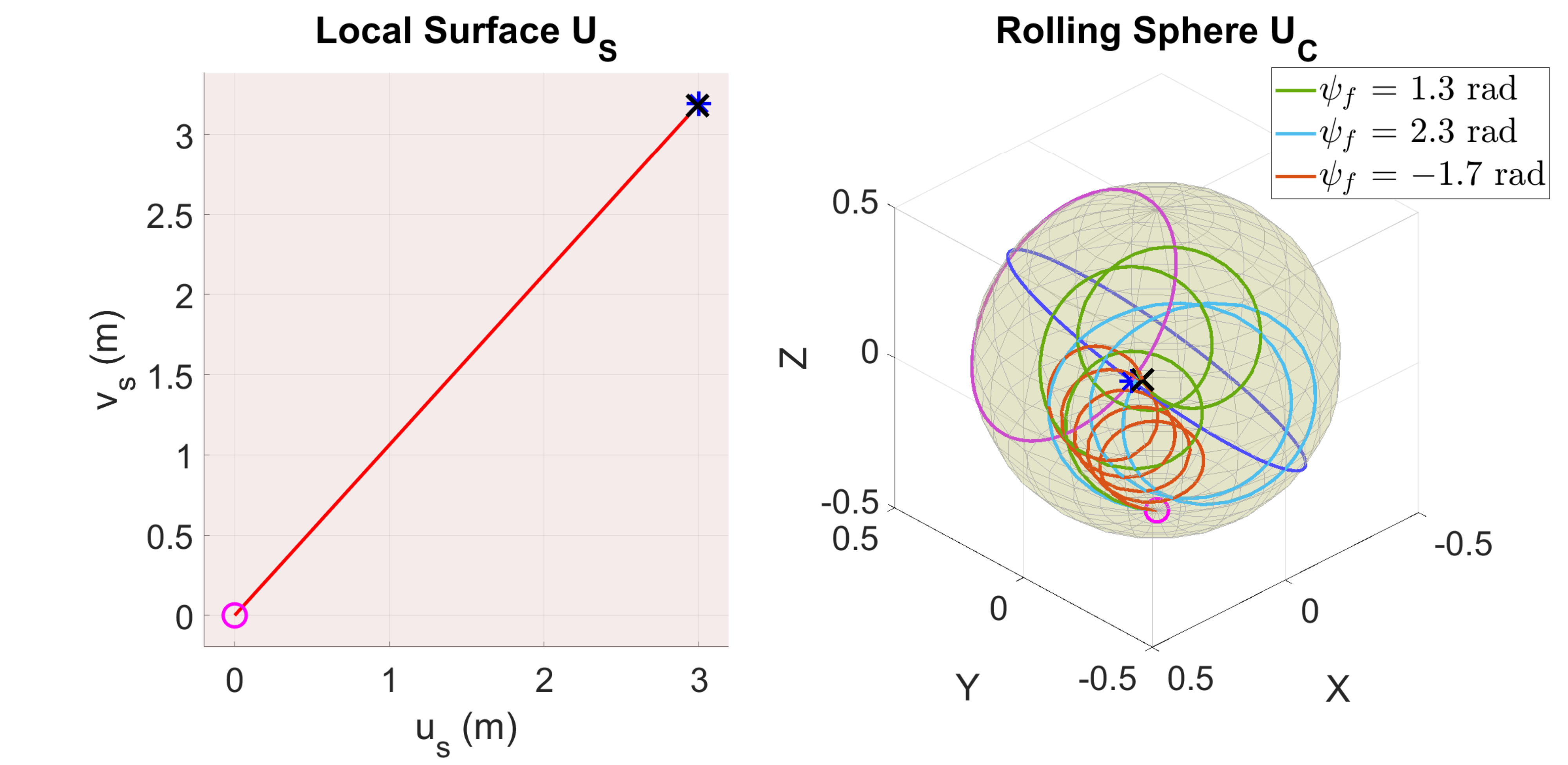}	\\
	\includegraphics[width=3.7 in, height= 3 in]{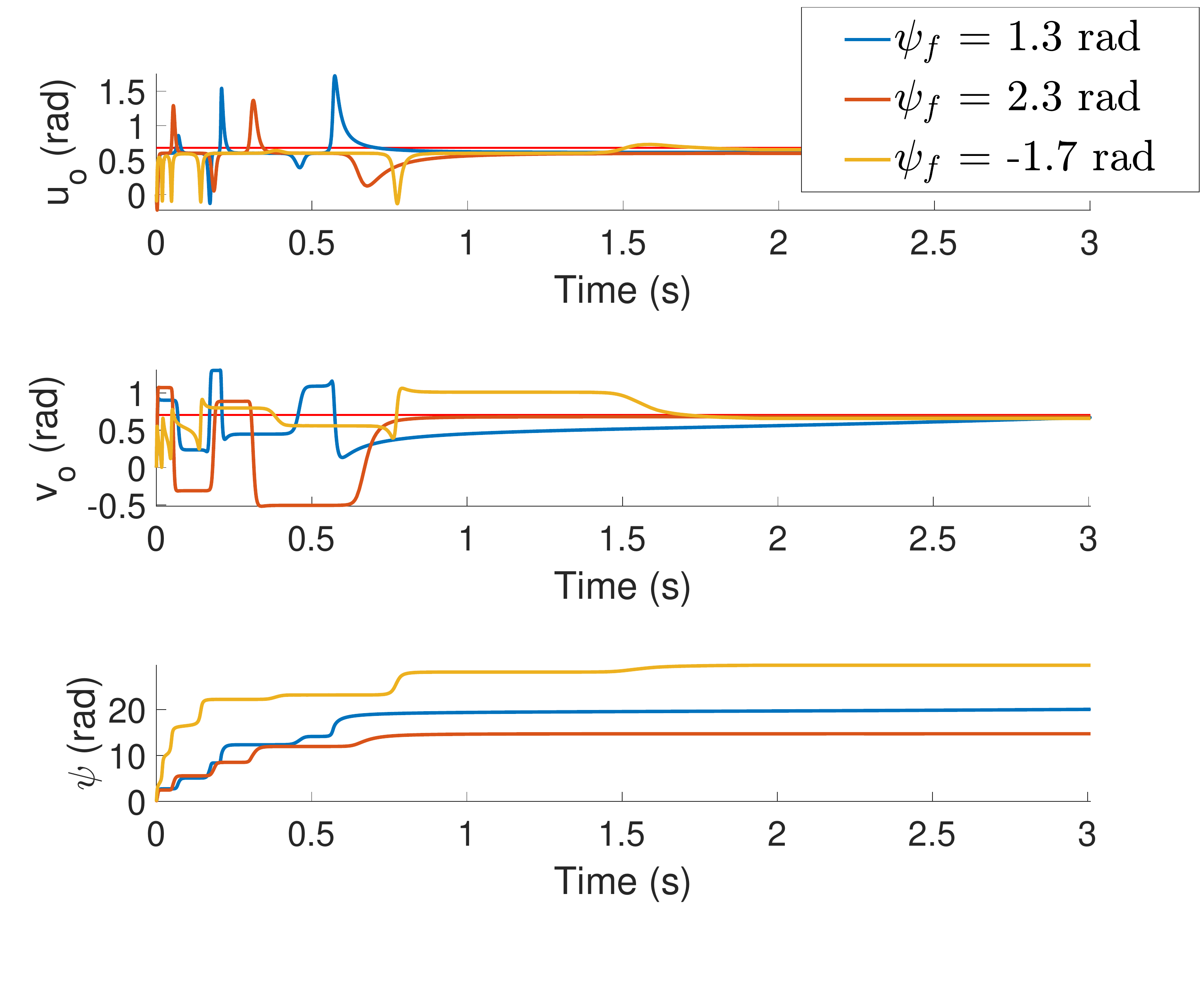}			
	\caption{Simulation results for different final spin angles $\psi_f=\{-1.7,1.3,2.3\}$.}\label{Fig:diifftrnetpsi}
\end{figure}

The angular velocities of the sphere collected by search algorithm are presented in Fig. \ref{Fig:Angular_Result1} but there are certain fast accelerations in the angular velocities. Note that the problem is considered in kinematics level without any slipping; Hence, arriving longer distances in a short amount of time results in faster velocities. However, we can use the aforementioned advantage in the derived kinematic model (\ref{EQ:LatestStateEquation}) where the time domain is separated from the kinematics by the rolling rate $\delta$. Thus, We alter the sphere angular velocities in two cases:

In the initial case, to decrease the overall velocity, the time span can be enlarged without affecting the achieved paths $\uvec{L}_o$ and $\uvec{L}_s$. For the considered example, we increase the time scaling constant $T$ (it is at $\delta$ in Eq. (\ref{Eq:DELTAEQuationtotal}) ) and simulation time $t_f$ to 160 s while the kinematic model is solved with the succeeded parameters in $\zeta'$ and $R_q+R_u$ from the algorithm. Clearly, the overall velocities decreased by expanding them in time as Fig. \ref{Fig:Beta}-b (dashed line velocities). However, there are fast accelerations (around 3 s and 21 s ) that makes it harder to be applied practically. To solve this issue, we keep the simulation time $t_f$ same as 160 s but a smooth function is chosen for $T$ as the second case study. With the known relation of $\omega^{o}_z= \delta k^{*}_g \overset{\Delta}{=}\delta \alpha_s$ \cite{Tafrishi2021DarKiF} and Eq. (\ref{Eq:DELTAEQuationtotal}), we define $T$ as
\begin{align}
T\overset{\Delta}{=}(c \cdot \omega_{s})/\alpha_s,
\end{align}
where $c=\left| v_{o,f} \cdot u'_o  \right|\cdot||\uvec{P}_f-\uvec{P}(t)||_2$ and $\omega_s$ is the desired smooth angular velocity. The desired smooth velocity $\omega_{s}$ is created with a symmetric time function as
\begin{align}
\omega_{s}= a \left( -\frac{140}{T_s^7}t^7+\frac{420}{T_s^6}t^6-\frac{420}{T_s^5}t^5+\frac{140}{T_s^4}t^4 \right),
\label{Eq:SmoothVelDesired}
\end{align}
where $a=12.91$ and $T_s=160$ are the time constant and the amplitude. Fig. \ref{Fig:Beta}-a shows the rest-to-rest function behavior of Eq. (\ref{Eq:SmoothVelDesired}). After running the same simulation with our defined $T$, we get the results in Fig. \ref{Fig:Beta}-b (solid lines velocities). It is clear that the second case has smooth angular velocities which are practical without large accelerations at the same time span of $t_f$. The angular displacement in Fig. \ref{Fig:Angular_Vel_com_pos1} also demonstrate how the desired angles of the sphere get expanded with the same convergence trend ($\uvec{L}_o$ trajectory is the same as Fig. \ref{Fig:Case1A1} ). Note that more complex functions can be defined for $T$ to converge with different angular velocities without affecting the trajectories of $\uvec{L}_o$ and $\uvec{L}_s$.

Next, we analyze the patterns of the created curves $\uvec{L}_o$ for different desired spin angles with the same final configuration. In here, the final configuration is considered  $\{u_{s,f}$, $v_{s,f}$, $u_{o,f}$, $v_{o,f}\}=\{3,3.2,0.6,0.7\}$ besides different desired spin angles $\psi_f=\{-1.7,1.3,2.3\}$. To have these different curves approximately on same configuration, the $ \epsilon_n$ and $\epsilon_r$ accuracies are chosen $\{0.045,0.05\}$, $\{0.048,0.034\}$ and $\{0.048,0.05\}$ for $\psi_f=\{-1.7,1.3,2.3\}$ cases, in the given order. Also, simulation time is set for $t_f=20$ s with $T=1$. Note that the rest of the initial conditions are the same as the previous case study.

Simulation results for different final spin angles are presented in Fig.~\ref{Fig:diifftrnetpsi}. As can be seen, the algorithm converges successfully to desired states with different trajectories. The average process time for these simulations was about 40 s. Note that the rise in the average process time is due to exceptional computation for $\psi_f=-1.7$ rad. This occurs depending on the accuracy and how near the desired $\bm\Psi_f$ angles are to $\bm\Psi_0$ which causes larger iterations (110 s process time).

Also, for the case of $\psi_f=-1.7$ rad, the curve creates multiple loops to reach the desired goal. As mentioned before, the reason is that some spin angles exhibit slower convergence by the designed approach. To achieve a faster convergence, we have to change the region of attraction for the desired configuration. This can be done by adding $+\frac{\pi}{2}$ to $v$-curve for the $\alpha_s(v_{o,f}+\frac{\pi}{2})$ arc-length-based input for this configuration.
Since we have already guarantied the convergence of the desired goal by the Theorem \ref{TheoremGlobalConvergence}, we can have regional shifts in the $v_{o,f}$ and $u_{o,f}$. Note that this is an interesting property of proposed algorithm since multiple solutions would be achievable for the same configuration with changing the time constant $T$, resulting to different regions of attraction on $u$- and $v$-curves.

\section{Conclusion}
In this paper, we presented a new approach for path planning of the spin-rolling sphere on a straight plane trajectory. The planning was mainly constructed by the introduced geometric control. In the beginning, the path planning problem is explained. Next, a virtual surface was designed to produce arc-length-based control inputs for the fully-actuated transformed model in \cite{Tafrishi2021DarKiF}. The problem is imagined as manipulating a flexible rope-like curve on the sphere with a constant length. Thus, we proposed an iterative algorithm to tune the traversed curve toward desired configurations. The achieved simulations clarified the performance of the proposed planning approach.

The proposed algorithm produces smooth trajectories of the spin-rolling sphere. In this connection, it should be noted that the feedback-based planning algorithms \cite{date2004simultaneous,Oriolo2005Feedback, das2004exponential, mukherjee2002feedback} results in piece-wise smooth trajectories. Also, our approach does not require decomposition of the planning strategy into motion steps as featured in geometric phase shift methods \cite{planningli1990,mukherjee2002feedback,date2004simultaneous,morinaga2014motion}. Another advantage is that the resulting angular velocities of the spin-rolling sphere can shaped to different desired convergence rates.

\bibliography{cas-refs}

\begin{thebibliography}{10}
\expandafter\ifx\csname url\endcsname\relax
  \def\url#1{\texttt{#1}}\fi
\expandafter\ifx\csname urlprefix\endcsname\relax\def\urlprefix{URL }\fi
\expandafter\ifx\csname href\endcsname\relax
  \def\href#1#2{#2} \def\path#1{#1}\fi

\bibitem{kiss2002motion}
B.~Kiss, J.~L{\'e}vine, B.~Lantos, On motion planning for robotic manipulation
  with permanent rolling contacts, Int. J. Robot. Res. 21~(5-6) (2002)
  443--461.

\bibitem{cui_sun_dai_2017}
L.~Cui, J.~Sun, J.~Dai, In-hand forward and inverse kinematics with rolling
  contact, Robotica 35~(12) (2017) 2381--2399.

\bibitem{SpinrollMechIROS2020}
S.~Yuan, L.~Shao, C.~L. Yako, A.~Gruebele, J.~K. Salisbury, Design and control
  of roller grasper v2 for in-hand manipulation, in: IEEE/RSJ International
  Conference on Intelligent Robots and Systems (IROS), 2020, pp. 9151--9158.

\bibitem{RollRollerRobotThesis2014}
S.~A. Tafrishi, {"RollRoller"} novel spherical mobile robot basic dynamical
  analysis and motion simulations, Master's thesis, University of Sheffield,
  Sheffield, UK (2014).

\bibitem{Tafrishi2019}
S.~A. Tafrishi, M.~Svinin, E.~Esmaeilzadeh, M.~Yamamoto, Design, modeling, and
  motion analysis of a novel fluid actuated spherical rolling robot, ASME J.
  Mech. Robot. 11~(4) (2019) 041010.

\bibitem{ishikawa2011volvot}
M.~Ishikawa, R.~Kitayoshi, T.~Sugie, Volvot: A spherical mobile robot with
  eccentric twin rotors, in: Proc. IEEE Int. Conf. Robot. Biomimetics, 2011,
  pp. 1462--1467.

\bibitem{karavaev2020spherical}
Y.~Karavaev, I.~Mamaev, A.~Kilin, E.~Pivovarova, Spherical rolling robots:
  Different designs and control algorithms, in: Robots in Human Life: Proc. of
  the 23rd Int. Conf. on Climbing and Walking Robots and the Support
  Technologies for Mobile Machines (CLAWAR 2020, Moscow, Aug 24--26 2020),
  2020, pp. 195--202.

\bibitem{borisov2012control}
A.~Borisov, A.~Kilin, I.~Mamaev, How to control chaplygin sphere using rotors,
  Regular and Chaotic Dynamics 17~(3) (2012) 258--272.

\bibitem{borisov2013control}
A.~Borisov, A.~Kilin, I.~Mamaev, How to control the chaplygin ball using
  rotors. ii, Regular and Chaotic Dynamics 18~(1) (2013) 144--158.

\bibitem{svinin2013dynamic}
M.~Svinin, A.~Morinaga, M.~Yamamoto, On the dynamic model and motion planning
  for a spherical rolling robot actuated by orthogonal internal rotors, Regular
  and Chaotic Dynamics 18~(1) (2013) 126--143.

\bibitem{fankhauser2010modeling}
P.~Fankhauser, C.~Gwerder, Modeling and control of a ballbot, {B.S.} thesis,
  Eidgen{\"o}ssische Technische Hochschule Z{\"u}rich (2010).

\bibitem{spinjohnson2018fuzzy}
J.~Johnson, R.~Senthilnathan, M.~Negi, R.~G. Patel, A.~Bhattacherjee, A fuzzy
  logic-in-loop control for a novel reduced height ballbot prototype, Procedia
  Comput. Sci. 133 (2018) 960--967.

\bibitem{sumer2008rolling}
B.~S{\"u}mer, M.~Sitti, Rolling and spinning friction characterization of fine
  particles using lateral force microscopy based contact pushing, J. Adhes Sci.
  Technol. 22~(5-6) (2008) 481--506.

\bibitem{diller2013micro}
E.~Diller, M.~Sitti, et~al., Micro-scale mobile robotics, Found. Trends Robot.,
  2~(3) (2013) 143--259.

\bibitem{fernandez2017three}
A.~Fern{\'a}ndez-Pacheco, R.~Streubel, O.~Fruchart, R.~Hertel, P.~Fischer,
  R.~P. Cowburn, Three-dimensional nanomagnetism, Nat. Commun. 8 (2017) 15756.

\bibitem{jurdjevic1993geometry}
V.~Jurdjevic, The geometry of the plate-ball problem, Arch. Ratio. Mech. Anal.
  124~(4) (1993) 305--328.

\bibitem{marigo2000rolling}
A.~Marigo, A.~Bicchi, Rolling bodies with regular surface: Controllability
  theory and applications, {IEEE} Trans. Autom. Control 45~(9) (2000)
  1586--1599.

\bibitem{Tafrishi2021DarKiF}
S.~A. Tafrishi, M.~Svinin, M.~Yamamoto, Darboux-frame-based parametrization for
  a spin-rolling sphere on a plane: A nonlinear transformation of underactuated
  system to fully-actuated model, Mechanism and Machine Theory 164 (2021)
  104415.

\bibitem{sankar1996velocity}
N.~Sankar, V.~Kumar, X.~Yun, Velocity and acceleration analysis of contact
  between three-dimensional rigid bodies.

\bibitem{Montana1988}
D.~J. Montana, The kinematics of contact and grasp, Int. J. Robot. Res. 7~(3)
  (1988) 17--32.

\bibitem{CuiDarboux2010}
L.~Cui, J.~S. Dai, A darboux-frame-based formulation of spin-rolling motion of
  rigid objects with point contact, IEEE Trans. Robot. 26~(2) (2010) 383--388.

\bibitem{bizyaev2019different}
I.~Bizyaev, A.~Borisov, I.~Mamaev, Different models of rolling for a robot ball
  on a plane as a generalization of the chaplygin ball problem, Regular and
  Chaotic Dynamics 24~(5) (2019) 560--582.

\bibitem{woodruff2019second}
Z.~Woodruff, K.~Lynch, Second-order contact kinematics between
  three-dimensional rigid bodies, Journal of Applied Mechanics 86~(8).

\bibitem{date2004simultaneous}
H.~Date, M.~Sampei, M.~Ishikawa, M.~Koga, Simultaneous control of position and
  orientation for ball-plate manipulation problem based on time-state control
  form, IEEE Trans. Robot. 20~(3) (2004) 465--480.

\bibitem{Oriolo2005Feedback}
G.~{Oriolo}, M.~{Vendittelli}, A framework for the stabilization of general
  nonholonomic systems with an application to the plate-ball mechanism, IEEE
  Trans. Robot. 21~(2) (2005) 162--175.

\bibitem{das2004exponential}
T.~Das, R.~Mukherjee, Exponential stabilization of the rolling sphere,
  Automatica 40~(11) (2004) 1877--1889.

\bibitem{alouges2010motion}
F.~Alouges, Y.~Chitour, R.~Long, A motion-planning algorithm for the
  rolling-body problem, IEEE Trans. Robot. 5~(26) (2010) 827--836.

\bibitem{mukherjee2002feedback}
R.~Mukherjee, T.~Das, Feedback stabilization of a spherical mobile robot, in:
  Proc. IEEE/RSJ Int. Conf. on Intell. Robots Sys., Vol.~3, 2002, pp.
  2154--2162.

\bibitem{woodruff2020motion}
Z.~Woodruff, S.~Ren, K.~Lynch, Motion planning and feedback control of rolling
  bodies, IEEE Access 8 (2020) 31780--31791.

\bibitem{planningli1990}
Z.~Li, J.~Canny, Motion of two rigid bodies with rolling constraint, IEEE
  Trans. Robot. Autom. 6~(1) (1990) 62--72.

\bibitem{mukherjee2002motion}
R.~Mukherjee, M.~A. Minor, J.~T. Pukrushpan, Motion planning for a spherical
  mobile robot: Revisiting the classical ball-plate problem, ASME Journal of
  Dynamic Systems, Measurement, and Control 124~(4) (2002) 502--511.

\bibitem{svinin2008motion}
M.~Svinin, S.~Hosoe, Motion planning algorithms for a rolling sphere with
  limited contact area, IEEE Trans. Robot. 24~(3) (2008) 612--625.

\bibitem{kilin2015spherical}
A.~Kilin, E.~Pivovarova, T.~Ivanova, Spherical robot of combined type: Dynamics
  and control, Regular and chaotic dynamics 20~(6) (2015) 716--728.

\bibitem{bai2018dynamic}
Y.~Bai, M.~Svinin, M.~Yamamoto, Dynamics-based motion planning for a
  pendulum-actuated spherical rolling robot, Regular and Chaotic Dynamics
  23~(4) (2018) 243--259.

\bibitem{arthurs1986hammersley}
A.~Arthurs, G.~Walsh, On hammersley's minimum problem for a rolling sphere, in:
  Math. Proc. Cambridge Philos. Soc., Vol.~99, Cambridge University Press,
  1986, pp. 529--534.

\bibitem{sachkov2010}
Y.~Sachkov, Maxwell strata and symmetries in the problem of optimal rolling of
  a sphere over a plane, Sb. Math 201~(7) (2010) 1029--1051.

\bibitem{mashtakov2011}
A.~Mashtakov, Y.~Sachkov, Extremal trajectories and the asymptotics of the
  maxwell time in the problem of the optimal rolling of a sphere on a plane,
  Sb. Math 202~(9) (2011) 1347--1371.

\bibitem{morinaga2014motion}
A.~Morinaga, M.~Svinin, M.~Yamamoto, A motion planning strategy for a spherical
  rolling robot driven by two internal rotors, IEEE Trans. Robot. 30~(4) (2014)
  993--1002.

\bibitem{beschatnyi2014optimal}
I.~Beschatnyi, The optimal rolling of a sphere, with twisting but without
  slipping, Mat. Sb. 205~(2) (2014) 157.

\bibitem{cui2020sliding}
L.~Cui, J.~Dai, Sliding-rolling Contact and In-hand Manipulation, World
  Scientific, 2020.

\bibitem{DIfgeometry1976}
M.~P. do~Carmo, Differential Geometry of Curves and Surfaces, 2nd Edition,
  Prentice-Hall, 1976.

\bibitem{Riemannian2002}
E.~Cartan, Riemannian Geometry in an Orthogonal Frame, 1st Edition, World
  Scientific Pub Co Inc, 2002.

\bibitem{tapp2016differential}
K.~Tapp, Differential geometry of curves and surfaces, Springer, New York,
  2016.

\bibitem{zangwill1969nonlinear}
W.~Zangwill, Nonlinear programming: a unified approach, Vol. 196, Prentice-Hall
  Englewood Cliffs, NJ, 1969.

\bibitem{amirhorde_2021}
A.~Tafrishi, M.~Svinin, M.~Yamamoto, Y.~Hirata,
  \href{{https://www.youtube.com/watch?v=7JK3VtT0aqs}}{Path planning of
  spin-rolling sphere on a plane} (November 2021).
\newline\urlprefix\url{{https://www.youtube.com/watch?v=7JK3VtT0aqs}}

\end{thebibliography}

\appendix
\section{Ball-Plate System Preliminaries}
\label{Basicballplate}

\begin{figure}[t!]
	\centering
	\includegraphics[width=3.6 in]{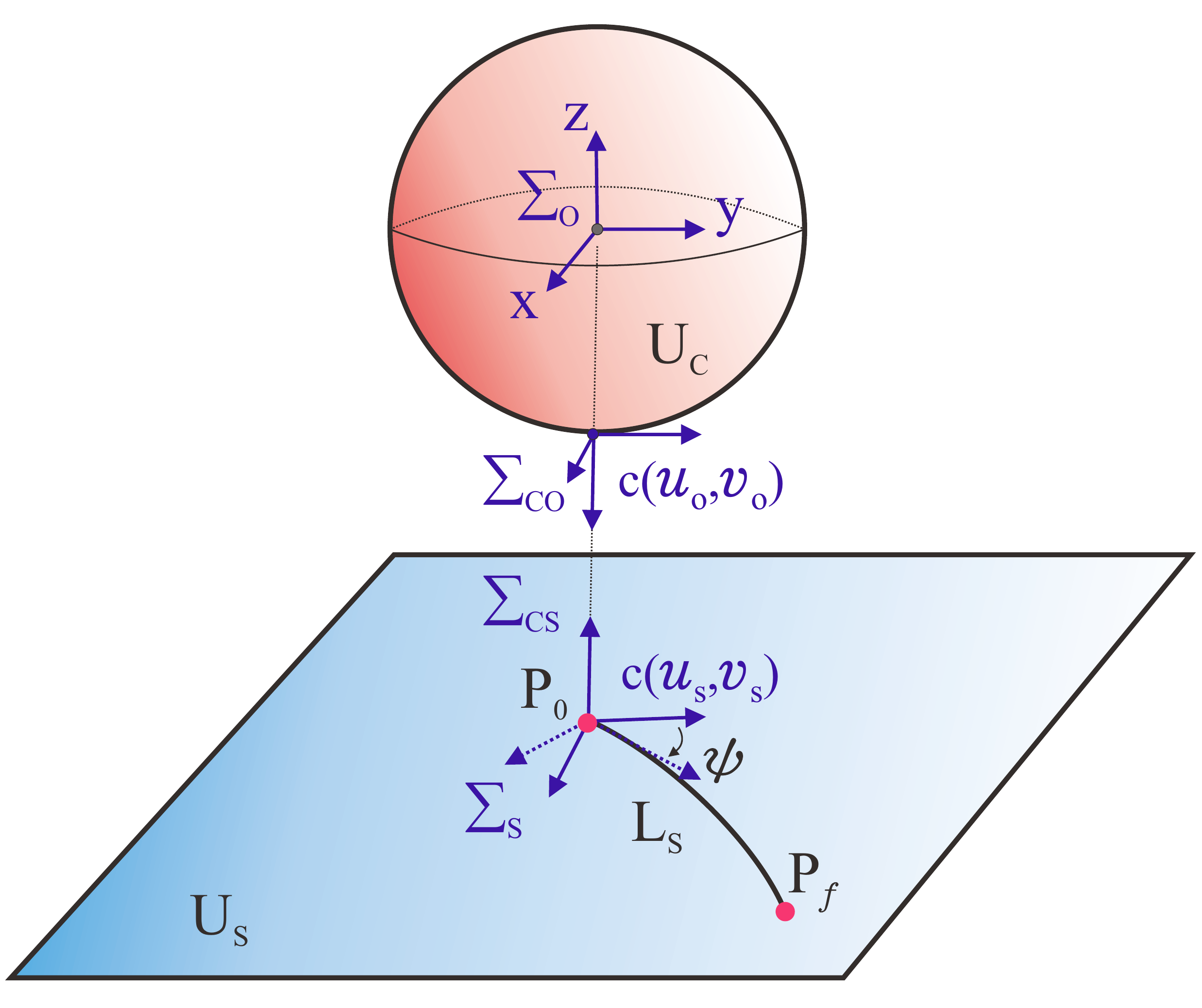}	
	\caption{Kinematic model of rotating sphere. Note: $\psi$ is the spin angle between sphere and plane surfaces.}\label{Fig:PlanningGeneralObject}
\end{figure}
Fig. \ref{Fig:PlanningGeneralObject} illustrates the rotating object and surface coordinates in the ball-plate system. Here, we have $\Sigma_{o}$ and $\Sigma_{s}$ as the fixed frames on the rolling object (red sphere) and the plane.
The frame $\Sigma_{s}$ is fixed relative to other coordinate frames. It is assumed that the sphere with radius $R_o$ is rotating with no sliding constraint. The local coordinate systems for the sphere and plane are considered
\begin{equation}
\begin{split}
&f_{o}: U_C \rightarrow \mathbb{R}^3
: c(u_{o},v_{o})\mapsto (-R_{o}\sin{u_{o}}\cos{v_{o}},R_{o}\sin{v_{o}},-R_{o}\cos{u_{o}}\cos{v_{o}}),\\
&f_{s}: U_S \rightarrow \mathbb{R}^3
: c(u_{s},v_{s}) \mapsto (u_{s},v_{s},0),
\end{split}
\label{Eq:Coorinatequationsphereonplane}
\end{equation}
where $c(u_{o},v_{o}) \in [-\pi,\pi]$ and $c(u_{s},v_{s})$ are contact parameters of the sphere and plane.
For the sphere we have \cite{planningli1990}:
\begin{equation}
k^{o}_{nu}=k^{o}_{nv}=1/R_o,\;\tau^{o}_{gu}=\tau^{o}_{gv}=0,\;k^{o}_{gu}=\tan(v_{o})/R_{o},\;k^{o}_{gv}=0,
\label{Eq:detailMontanaCOonplane}
\end{equation}
where $k^{o}_{nu}$, $k^{o}_{nv}$, $\tau^{o}_{gu}$, $\tau^{o}_{gv}$, $k^{o}_{gu}$, $k^{o}_{gv}$ are the normal curvature, geodesic torsion and geodesic curvature of the rotating body respect to $u_o$ and $v_o$ principle angle.
The curvature terms of the plane surface $U_S$ are
\begin{align}
\begin{split}
&k^{s}_{nu}=k^{s}_{nv}=\tau^{s}_{gu}=\tau^{s}_{gv}=k^{s}_{gu}=k^{s}_{gv}=0,
\label{Eq:MontanaCSNN}
\end{split}
\end{align}
where $k^{s}_{nu}$, $k^{s}_{nv}$, $\tau^{s}_{gu}$, $\tau^{s}_{gv}$, $k^{s}_{gu}$, $k^{s}_{gv}$ are the normal curvature, geodesic torsion and the geodesic curvature of the fixed surface (plane) respect to $u_s$ and $v_s$ principle angles.

\section{Geodesic Torsion Design of the Virtual Surface }
\label{VirtualSurfaceGeodesicTorsionProof}
The rotating object and plane do not have the geodesic torsion, $\tau^*_g=-\beta_{s}$. It makes the kinematic model (\ref{EQ:LatestStateEquation}) uncontrollable \cite{Tafrishi2021DarKiF}. Thus, a helicoid virtual surface with similar curvature properties with rotating object (see Fig. \ref{Fig:Virtualsurface123}) is proposed as
\begin{align}
\begin{split}
&f_{v}: U_V \rightarrow R^3
: c(u_{v},v_{v}) 	\mapsto (-R_{v}\sin{u_{v}}\cos{v_{v}},R_{v}\sin{v_{v}}+R_tu_{v},-R_{v}\cos{u_{v}}\cos{v_{v}}),
\end{split}
\label{EQ:Virtualsurface}
\end{align}
where $R_{v}$ and $R_t$ are defined by main spherical and sum of spherical and torsion radii, respectively. The curvature properties can be obtained as
\begin{align}
k^{v}_n=\frac{1}{R_{v}},\; k^{v}_g=\frac{R_{v}\cos{v_{v}}\sin{v_{v}}}{R^2_{v}\cos^2{v_{v}}+R^2_t},  \tau^{v}_g=\frac{1}{R_{v}^2}(R_{v}^2\cos^2{v_{v}}+R^2_t)^{\frac{1}{2}}
\label{EQ:CurvaturepropertiesVirtual}
\end{align}
where $k^{v}_n,\;k^{v}_g,\;\tau^{v}_g$ are normal curvature, geodesic curvature and geodesic torsion. We here mainly care about $\tau^{v}_g(R_{v},R_t,v_{v})$ and $k^{v}_n(R_{v})$ to understand relation of geodesic torsion design. Thats why we related them with separate corresponding radii. To prove that this surface let us manipulate $\tau_g$ by using $R_t$, we consider $R_t=nR_{v}$ where $n>1$, which results
\begin{align}
\tau^{v}_g=\frac{1}{R_{v}}(\cos^2{v_{v}}+n^2)^{\frac{1}{2}}.
\label{EQ:TorsionVirtualProof}
\end{align}

\begin{thm}
	At an arbitrary point $P$ on a surface, geodesic torsion relation with normal curvature is defined as \cite{Riemannian2002}
	\begin{align}
	\tau_g=\frac{1}{2}(k_{nu}-k_{nv}) \sin 2\Omega
	\label{Eq:TheoremGeodesicNormal}
	\end{align}
	where $k_{nu}$ and $k_{nv}$ are the principle curvatures. Also, $\Omega$ is the counterclockwise angle from the direction of minimum curvature $k_{nu}$ on the tangent plane.	
	\label{Theorem:geodeisc}
\end{thm}
By relying on the Theorem \ref{Theorem:geodeisc}, it is clear that sphere tangent plane angle $\Omega$ is $\pi/4$. Now, we can do some algebraic operations on (\ref{Eq:TheoremGeodesicNormal}) with considering $k_{nu}=1/R_{v}$ and Eq. (\ref{EQ:TorsionVirtualProof}),
\begin{equation}
k_{nv}=1/R_{v}\left[1+(\cos^2{v_{v}}+n^2)^{\frac{1}{2}}\right]
\end{equation}
where $k_{nv}$ is the second principle curvature of the virtual surface. This supports our assumption that designed geodesic torsion is inverse of sphere radius with similar unity of normal curvature $R_t=n R_{v}$. We do this for sake of separating the radius of geodesic torsion and normal curvature in Eq. (\ref{Eq:TheCurvaturepropertymodeldesign}) as we use two separate surfaces (sphere and helicoid) in arc-length-based inputs.

\section{Phase I Directional Update}
\label{RegionalComparison}
\begin{figure}[t!]
	\centering	
		\includegraphics[width=2.6 in,height=2.3 in]{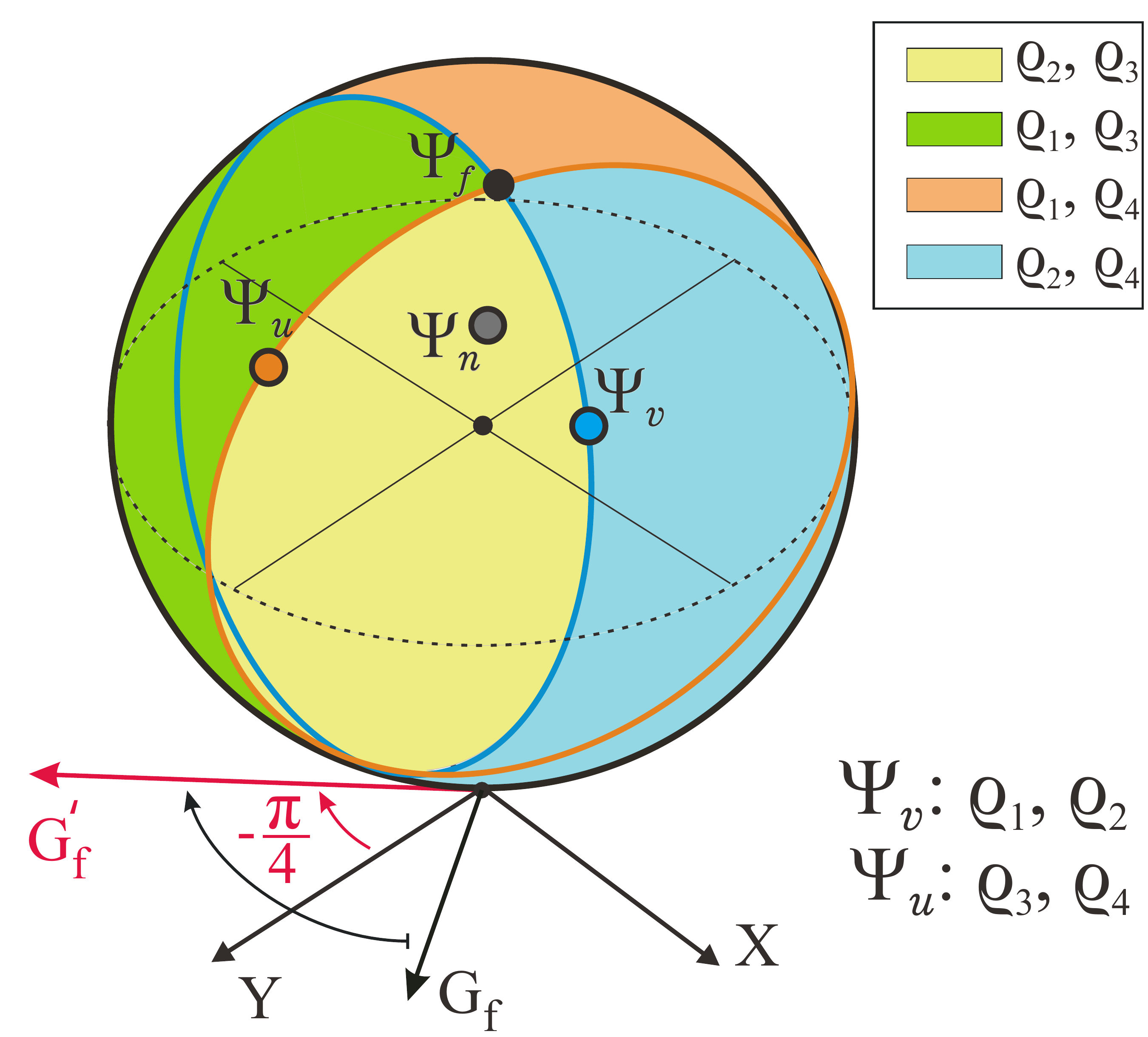}				
	\caption{The sectioned sphere based on the $\bm{\Psi}_v$ and $\bm{\Psi}_u$ cutting planes for directional updates of $\zeta_q$. Note that $\bm{\Psi}_v$ and $\bm{\Psi}_u$ are on $\varrho_2$ and $\varrho_3$ as an example.}\label{Fig:SectionizedTracking}
\end{figure}

We can separate the sphere regions to four by $c(u_{o,f},v_{o,f})$ cutting planes (see example in Fig. \ref{Fig:SectionizedTracking}) for the desired configuration where $\zeta_q$ is calculated for its directional updates.
Because we need to find directional updates in different $G_f$ angles on plane, we apply a spherical rotational transformation to ease our regional computation. To find corresponding local points' right location for the directional update from plane angle $G_f$ respect to $v_{s}$, we use an arbitrary local coordinate $c(u_o,v_o)$ rotation respect to $-\pi/4$ base angle, $G'_f=G_f-\pi/4$, where gives new coordinate $c(u^r_o,v^r_o)$ as
\begin{equation}
\begin{split}
& v^r_o= \sin^{-1}\big [-\sin G'_f \sin u_o \cos v_o+ \cos G'_f \sin v_o\big],\;u^r_o=\sin^{-1}\big [(\cos G'_f \sin u_o \cos v_o+ \sin G'_f \sin v_o)/\cos v^r_o]\\
\end{split}
\label{EQ:GoalRotationsSphere}
\end{equation}
Next, to avoid the numerical solution of following Eqs. (\ref{EQ:GoalRotationsSphere}), we present it in algebraic form depending on $c(u_o,v_o)$ location by
\begin{equation}
\begin{cases}
& \begin{cases}
&u^r_o \leftarrow -u^r_o, \\
&v^r_o \leftarrow \pi-v^r_o,
\end{cases} \;\;\textnormal{for}\;  \Big[\left(0 \leq |v_o| \leq \frac{\pi}{2}\; \& \;\frac{\pi}{2} \leq |u_o| \leq \pi \right) || \left(\frac{\pi}{2}<|v_o| \leq \pi\; \& \;0 \leq |u_o|<\frac{\pi}{2} \right) \Big] \\
& \begin{cases}
&u^r_o \leftarrow u^r_o, \\
&v^r_o \leftarrow  v^r_o,
\end{cases} \;\;\;\;\;\;\;\;\textnormal{for}\;  \Big[\left(0 < |v_o|< \frac{\pi}{2}\; \& \;0<|u_o|<\frac{\pi}{2} \right) || \left(\frac{\pi}{2}<|v_o|<\pi\; \& \;\frac{\pi}{2}<|u_o|<\pi \right) \Big] \\
\end{cases}
\label{Eq:RegionalRotaionofpoints}
\end{equation}
The operations (\ref{EQ:GoalRotationsSphere})-(\ref{Eq:RegionalRotaionofpoints}) are applied to rotate $\bm{\Psi}_f$, $\bm{\Psi}_u$, $\bm{\Psi}_v$ and $\bm{\Psi}_n$ relative to angle $G_f$ to find them always in one $G'_f$ direction. Now, the sectioned sphere as Fig. \ref{Fig:SectionizedTracking} is obtained by cutting planes $\bm{\Psi}_v$ and $\bm{\Psi}_u$ where they create $\{\varrho_1,\varrho_2\}$ and $\{\varrho_3,\varrho_4\}$ regions, respectively. Note that for nearest point $\bm{\Psi}_n$ on $\uvec{L}_o$, the corresponding $\varrho$ are shown as $\varrho_{n1}$, $\varrho_{n2}$, $\varrho_{n3}$, $\varrho_{n4}$. We design our directional updates by comparing $\bm{\Psi}_n$ with $\bm{\Psi}_u$, $\bm{\Psi}_v$ as following computation
\alglanguage{pseudocode}
\begin{algorithmic}
	\State Calculate $G'_f$ depending on the $\{u_{s,f},v_{s,f}\}$
	\State Calculate rotated coordinates of $\bm{\Psi}_f$, $\bm{\Psi}_u$, $\bm{\Psi}_v$ and $\bm{\Psi}_n$ by (\ref{Eq:RegionalRotaionofpoints})
	\State Calculate $Q^{zy}_{f}$, $Q^{zx}_{f}$, $Q^{zy}_{n}$ and $Q^{zx}_{n}$ according to (\ref{Eq:TheQCOFormulas})
	\If {$ \varrho_{2} = 1\; \& \; \varrho_{4}= 1 $} 	 \Comment{Exceptional regional updates}
	\If {only $ \varrho_{n2} = 1 $}
	\State $\zeta_q(k)\leftarrow +\zeta_q(k-1)$
	\ElsIf {$\left( \varrho_{n2} = 1\; \& \; \varrho_{n4}= 1\right)$ $||$ $($only $\varrho_{n4}= 1)$}
	\If {$v^r_{o,f} \geq 0$}
	\If {$u^r_{o,f} \geq 0$}
	\State $\zeta_q(k)\leftarrow +\zeta_q(k-1)$
	\Else
	\State $\zeta_q(k)\leftarrow -\zeta_q(k-1)$
	\EndIf
	\Else
	\State $\zeta_q(k)\leftarrow -\zeta_q(k-1)$
	\EndIf
	
	\EndIf
	\ElsIf {$ \left(\varrho_{n2} = 1 \;\& \;\varrho_{n3}= 1\right) || \left( \varrho_{3} = 1 \;\& \;\varrho_{n2}= 1 \;\& \;\varrho_{n4}= 1 \right)$}	
	\If {$u^r_{o,f} \geq 0$}
	\State $\zeta_q(k)\leftarrow +\zeta_q(k-1)$
	\Else
	\State $\zeta_q(k)\leftarrow -\zeta_q(k-1)$
	\EndIf
	\ElsIf {$ \varrho_{1}=\varrho_{2}=\varrho_{3}=\varrho_{4} = 0 $} 	
	\If {$v^r_{o,f} \geq 0$}
	\State $\zeta_q(k)\leftarrow -\zeta_q(k-1)$
	\Else
	\State $\zeta_q(k)\leftarrow +\zeta_q(k-1)$
	\EndIf
	\Else \Comment{Normal regional updates}
	\If {only $\left( \varrho_{n1} = 1\right)$ $||$ $\left(\varrho_{n4} = 1\right)$} 	
	\If {$v^r_{o,f} \geq 0$}
	\State $\zeta_q(k)\leftarrow +\zeta_q(k-1)$
	\Else
	\State $\zeta_q(k)\leftarrow -\zeta_q(k-1)$
	\EndIf
	\ElsIf {only $\left( \varrho_{n2} = 1 \right)$ $||$ $\left(\varrho_{n3} = 1 \right)$}
	\If {$v^r_{o,f} \geq 0$}
	\State $\zeta_q(k)\leftarrow -\zeta_q(k-1)$
	\Else
	\State $\zeta_q(k)\leftarrow +\zeta_q(k-1)$
	\EndIf
	\EndIf
	
	\EndIf	
\end{algorithmic}
where $\varrho_{ni}$ presents the nearest point $\bm{\Psi}_n$ location on $i$-th section of the spherical surface ($i\in[1,4]$) with respect to $\bm{\Psi}_f$. Also, $\varrho_{ni}=1$ means the following argument is true. Note that this complexity in sectioned spherical surface, as shown in Fig.~\ref{Fig:SectionizedTracking}, is due to rolling surface that creates different signs of updates of $\zeta_q(k)$ in (\ref{Eq:UpdateofZetaq}) based on $\{\bm{\Psi}_0,\bm{\Psi}_f, \bm{\Psi}_n\}$. In additions, the normal updates in the algorithm find $\bm\Psi_n$ location on the spherical surface and compare it to the location of desired goal $\bm\Psi_f$ to give correct sign update for $\zeta_q(k)$. However, exceptional updates are designed to exist from trajectory loops of $\uvec{L}_o$ for $\bm\Psi_n$ and bring the $\uvec{L}_o$ curve toward the location of $\bm\Psi_0$.
%

\section{Proof of Closed Set}
\label{ProofofClosedSet}
Let $\uvec{x}_k\rightarrow\uvec{x}_0$ and $\uvec{d}_k \rightarrow \uvec{d}_0$ be the convergence to desired values as $k\rightarrow \infty $ via sequences of $\{\uvec{x}_k  \}^{\infty}_{k=1}$ and $\{\uvec{d}_k  \}^{\infty}_{k=1}$. Also, suppose $\{\uvec{y}_k  \}^{\infty}_{k=1}$ is a sequence where $\uvec{y} \in \Gamma(\uvec{x}_k,\uvec{d}_k)$ for all $k$ while it has convergence of $\uvec{y}_k\rightarrow\uvec{y}_0$ as $k\rightarrow \infty$. We want to illustrate that $\uvec{y}_0 \in \Gamma(\uvec{x}_0,\uvec{d}_0) $ for having a closed set.

In each iteration $k$, $\uvec{y}_k=\uvec{x}_k+\uvec{h}_k \uvec{d}_k$ for $\uvec{h}_k>0$. Thus, by knowing fact that $||\uvec{d}_k||=[1\;..\;1]^T$, there is
\begin{equation*}\small{
	\begin{split}
	&\uvec{h}_k=||\uvec{y}_k-\uvec{x}_k||= \left[\begin{array}{cccc}
	||\uvec{y}_k(1)-\uvec{x}_k(1)|| &...\\
	...&   ||\uvec{y}_k(n)-\uvec{x}_k(n)||
	\end{array}\right]  \rightarrow  \uvec{h}^*=||\uvec{y}_0-\uvec{x}_0||,
	\end{split}  }
\end{equation*}
hence it implies that $\uvec{y}_0 = \uvec{x}_0 + \uvec{h}^* \uvec{d}_0 $.

Now, to show $\uvec{y}_0$ minimizes the $f$ along the $\uvec{x}_0+\uvec{h}\uvec{d}_0$, we know for each k and $\uvec{h}$, $0<\uvec{h}<\infty$, there is
\begin{equation*}
f(\uvec{y}_k)\leq f(\uvec{x}_k+\uvec{h}\uvec{d}_k).
\end{equation*}
Therefore, by continuity of $f$, $k\rightarrow \infty$ leads to $f(\uvec{y}_0)\leq f(\uvec{x}_0+\uvec{h}\uvec{d}_0)$ for all $\uvec{h}$ in which shows
\begin{equation*}
f(\uvec{y}_k)\leq \min { \;f(\uvec{x}_k+\uvec{h}\uvec{d}_k) }.
\end{equation*}
This proves the definition of $\uvec{y}_0 \in \Gamma(\uvec{x}_0,\uvec{d}_0) $ for having a closed set.

\end{document}